\documentclass[dvipsnames,sigconf=true, nonacm=true, review=false, anonymous = false]{acmart}
\AtBeginDocument{%
  }

\usepackage{float}
\usepackage{multirow}
\usepackage{adjustbox}
\usepackage{tcolorbox}
\usepackage{soul}               
\usepackage{booktabs}           
\usepackage{colortbl}           
\graphicspath{{pics/}}          
\usepackage{tikz}
\usetikzlibrary{shapes.geometric}
\usetikzlibrary{positioning,fit,shapes,shadows,arrows,calc}
\usetikzlibrary{intersections}
\usepackage{etoolbox}           
\usepackage{amsmath}            
\usepackage{amsfonts}           
\usepackage{amsthm}             
\usepackage{xspace}             
\usepackage{dsfont}             
\usepackage{rotating}           
\usepackage{bm}                 
\usepackage{subcaption}         
\usepackage{hyperref}           
\usepackage{mathtools}          
\usepackage{url}                
\usepackage[noabbrev,nameinlink]{cleveref} 
\usepackage{bbm}

\usepackage{pgfplots}  
\usepackage{tikz}      
\usepackage{pgfplotstable}  
\pgfplotsset{compat=1.15}
\pgfdeclarelayer{background}
\pgfdeclarelayer{foreground}
\pgfsetlayers{background,main,foreground}
\usepackage[skip=7pt]{caption}

\usepackage[algo2e,ruled,vlined,linesnumbered]{algorithm2e}

\usepackage{cleveref}

\tikzstyle{activity}=[align=center, rectangle, draw=black, rounded corners, text width=8em, fill=white, drop shadow]
\tikzstyle{data}=[align=center, rectangle, draw=black, fill=black!10, text width=8em, drop shadow]
\tikzstyle{internal}=[rectangle, draw=black!60, thick, align=left, text width=8em, minimum height=.75cm, text=black!60]
\tikzstyle{myarrow}=[-latex, thick]

\DeclareMathOperator*{\argmax}{arg\,max}

\DeclareMathOperator{\Bin}{Bin}
\newcommand{\R}{\mathbb{R}}
\newcommand{\N}{\mathbb{N}}

\definecolor{pastelred}{rgb}{1.0, 0.6, 0.6}
\definecolor{pastelgreen}{rgb}{0.6, 1.0, 0.6}
\definecolor{pastelblue}{rgb}{0.6, 0.6, 1.0}
\definecolor{pastelcyan}{rgb}{0.6, 1.0, 1.0}
\definecolor{pastelmagenta}{rgb}{1.0, 0.6, 1.0}
\definecolor{pastelyellow}{rgb}{1.0, 1.0, 0.6}
\definecolor{pastelorange}{rgb}{1.0, 0.8, 0.6}

\newcommand{\irace}{\textsc{irace}\xspace}
\newcommand{\onell}{(1+($\lambda$,$\lambda$))-GA\xspace}
\newcommand{\onemax}{\textsc{OneMax}\xspace}
\newcommand{\leadingones}{\textsc{LeadingOnes}\xspace}
\newcommand{\onemaxdac}{\textsc{OneMax-DAC}\xspace}
\newcommand{\lbd}{$\lambda$\xspace}
\newcommand{\K}{\mathcal{K}} %

\DeclareMathOperator{\flip}{flip}
\DeclareMathOperator{\cross}{cross}

\newcommand{\assign}{\leftarrow}

\begin{document}

\title[On the Importance of Reward Design in Reinforcement Learning-based Dynamic Algorithm Configuration]{On the Importance of Reward Design in Reinforcement Learning-based Dynamic Algorithm Configuration: \\A Case Study on OneMax with (1+($\lambda$,$\lambda$))-GA}

\author{Tai Nguyen}
\orcid{0009-0004-7707-2069}
\affiliation{
  \institution{University of St Andrews}
  \city{St Andrews}
  \country{United Kingdom}
}
\affiliation{
\institution{Sorbonne Universit\'e, CNRS, LIP6}
\city{Paris}
\country{France}
}

\author{Phong Le}
\orcid{0009-0000-0749-9519}
\affiliation{
  \institution{University of St Andrews}
  \city{St Andrews}
  \country{United Kingdom}}

\author{André Biedenkapp}
\orcid{0000-0002-8703-8559}
\affiliation{
  \institution{University of Freiburg}
  \city{Freiburg}
  \country{Germany}}

\author{Carola Doerr}
\orcid{0000-0002-4981-3227}
\affiliation{
  \institution{Sorbonne Universit\'e, CNRS, LIP6}
  \city{Paris}
  \country{France}}

\author{Nguyen Dang}
\orcid{0000-0002-2693-6953}
\affiliation{
  \institution{University of St Andrews}
  \city{St Andrews}
  \country{United Kingdom}}

\renewcommand{\shortauthors}{Nguyen et al.}

\begin{abstract}
  Dynamic Algorithm Configuration (DAC) has garnered significant attention in recent years, particularly in the prevalence of machine learning and deep learning algorithms. Numerous studies have leveraged the robustness of decision-making in Reinforcement Learning (RL) to address the optimization challenges associated with algorithm configuration. However, making an RL agent work properly is a non-trivial task, especially in reward design, which necessitates a substantial amount of handcrafted knowledge based on domain expertise. In this work, we study the importance of reward design in the context of DAC via a case study on controlling the population size of the $(1+(\lambda,\lambda))$-GA optimizing OneMax. We observed that a poorly designed reward can hinder the RL agent's ability to learn an optimal policy because of a lack of exploration, leading to both scalability and learning divergence issues. To address those challenges, we propose the application of a reward shaping mechanism to facilitate enhanced exploration of the environment by the RL agent. Our work not only demonstrates the ability of RL in dynamically configuring the $(1+(\lambda,\lambda))$-GA, but also confirms the advantages of reward shaping in the scalability of RL agents across various sizes of OneMax problems.
\end{abstract}


\maketitle

\sloppy

\section{Introduction}
\label{sec:introduction}

Evolutionary algorithms (EAs) are well-established optimization approaches, used to solve broad range of problems in various application domains every day. A key factor of EAs' wide adoption in practice is the possibility to adjust their search behavior to very different problem characteristics. To benefit from this versatility, the exposed parameters of an EA need to be suitably configured. Since this tuning task requires substantial problem expertise if done by hand, researchers have developed automated algorithm configuration (AC) tools to support the user by automating this process~\cite{hutter-jair09a}. AC tools such as \irace~\cite{lopez2016irace} and SMAC~\cite{SMAC3} are today quite well established and broadly used, especially in academic contexts, with undeniable success in various application domains~\cite{schede2022survey}. 

\emph{Dynamic algorithm configuration (DAC)} extends automated algorithm configuration by adapting parameters during optimization runtime rather than using fixed values throughout. While modern evolution strategies like CMA-ES~\cite{hansen2006cma} utilize control mechanisms based on current-run data, DAC aims to learn optimal parameter settings across multiple problem instances through transfer learning.
Initially explored in~\cite{lagoudakis2000algorithm} and then in~\cite{KeeAdaptiveGA,pettinger-gecco02b,aine-jasoc08a,sakurai-sitis10ab,KarafotiasSE12,BurkeGHKOOQ13,andersson-gecco16b,VermettenCMAdynAS,sharma2019deep}, the problem of learning control policies through a dedicated training process was formally introduced as DAC in~\cite{biedenkapp2020dynamic,adriaensen2022automated}. 

Given the large success of reinforcement learning (RL)~\cite{712192}  in similar settings where one wishes to control state-specific actions, such as in games \cite{mnih2013playing,silver2017mastering,gtsophy}, robotics \cite{lillicrap2015continuous,sac,loon}, or otherwise complex physical systems \cite{degrave-nature22a,droneracing}, it seems natural to address the DAC problem with RL approaches. In fact, the study~\cite{sharma2019deep} previously employed a double DQN method to determine the selection of mutation strategies in differential evolution. Recently, a multi-agent RL approach was used to control multiple parameters of a multi-objective evolutionary algorithm \cite{madac}.

Despite all successes, a number of recent studies also highlight the difficulty of solving DAC problems. 
Using theory-inspired benchmarks with a known ground truth, the analysis in~\cite{biedenkapp2022theory} revealed that a na\"ive application of RL to DAC settings can be fairly limited, with unfavorable performance in settings of merely moderate complexity. An alternative approach to address DAC problems via sequential algorithm configuration using \irace~\cite{lopez2016irace} was suggested in~\cite{chen2023using}. However, this approach requires substantial computational overhead, and its usefulness for more complex settings remains to be demonstrated. The so-called GPS strategy \cite{levine-neurips14}, used in~\cite{shala2020learning} to control the step-size of CMA-ES, requires itself a complex configuration process, causing significant overhead. 

\textbf{Our Contributions.} 
We revisit the DAC problem of configuring the $\lambda$ parameter of the \onell for optimizing \onemax instances, as introduced in~\cite{chen2023using}, but address with an RL approach. Specifically, we investigate DDQN~\cite{van2016deep}, a widely used deep RL algorithm in both general RL research and DAC applications. Our initial experiments in \Cref{sec:ddqn_for_onemax} reveal that a na\"ive implementation faces several challenges, including scalability issues across different problem sizes and significant divergence during learning.
 
 We show that these issues arise from a na\"ive reward function design (\Cref{sec:reward_scaling}). To address this, we propose a scaling mechanism to mitigate the impact of problem dimensionality on the reward function. Although effective for small- to medium-sized problems, this approach is insufficient for larger-scale problems.

We identify \emph{under-exploration} as the root cause of both scalability and divergence issues (\Cref{sec:under_exploration}) and show that the widely used $\epsilon$-greedy exploration~\cite{sutton1988learning} in RL is insufficient to solve this issue.

In~\Cref{sec:reward_shifting}, we resolve under-exploration and mitigate both scalability and divergence issues by adopting a simple method known as \textit{reward shifting} \cite{sun2022exploit}, which adds a bias term to the original reward. Additionally, we propose an adaptive mechanism to autmatically adjust the bias for reward shifting to avoid time-consuming manual tuning of this parameter.
Using this, our trained RL policies consistently outperform a well-known theory-derived policy for the same benchmark~\cite{doerr2018optimal} across all problem sizes studied. 
Finally, in~\Cref{sec:scalability_analysis}, we demonstrate that RL policies trained with our adaptive reward shifting can achieve a speed increase of several orders of magnitude to reach the performance of the theory-derived policy compared to the DAC method based on \irace proposed in~\cite{chen2023using}.

All source code and data are publicly available at~\cite{source}.

\section{Background}%
\label{sec:background}%
DAC problems are modelled as Markov Decision processes (MDPs) \cite{MDP}.
An MDP $\mathcal{M}$ is a tuple $(\mathcal{S}, \mathcal{A}, \mathcal{T}, \mathcal{R})$ with state space $\mathcal{S}$, action space $\mathcal{A}$,  transition function $\mathcal{T}\colon\mathcal{S}\times\mathcal{A}\times\mathcal{S}\to[0,1]$, and reward function $\mathcal{R}\colon\mathcal{S}\times\mathcal{A}\to\mathbb{R}$.
The transition function gives the probability of reaching a successor state $s'$ when playing action $a$ in the current state $s$, thus describing the dynamics of the system.
The reward function further indicates if such a transition between states is desirable or if it should be avoided and is crucial for learning processes that aim to learn policies that are able to solve the MDP. In order to describe instance-dependent dynamics and enable learning across multiple instances $i\sim\mathcal{I}$, DAC problems are described as contextual MDPs (cMDPs) \cite{hallak-corr15}.
Contextual MDPs extend the MDP formalism through the use of \emph{contextual information} that describes how rewards and transitions differ for different instances while sharing action and state spaces.
Consequently, a cMDP $\mathcal{M}=\{\mathcal{M}_i\}_{i\sim\mathcal{I}}$ is a collection of MDPs with shared state and action spaces, but with individual transition and reward functions.

In DAC, a state space represents the algorithm's behavior through its internal statistics during execution, providing necessary context. The action space encompasses all potential parameter configurations. While transition and reward functions are typically unknown and complex to approximate, RL has proven effective for DAC \cite{adriaensen2022automated}. During offline learning, an RL agent interacts with the algorithm being tuned across multiple episodes, each terminating at a goal state or step limit. The agent observes the current state $s_t$, selects action $a_t$, transitions to state $s_{t+1}$, and receives reward $r_{t+1}$. These interactions enable the agent to evaluate states and determine an optimal policy $\pi\colon\mathcal{S}\to\mathcal{A}$ maximizing expected rewards. Though some RL variants learn transition functions for direct policy search, modern approaches are typically model-free, focusing on learning state-action values $\mathcal{Q}\colon\mathcal{S}\times\mathcal{A}\to\mathbb{R}$.

$\mathcal{Q}$-learning \cite{watkins1992q}, one of the most widely adopted approaches, aims to learn a $\mathcal{Q}$-function that associates each state-action pair with its expected cumulative future reward when taking action $a$ in state $s$.
This function is learned through error correction principles.
For a given state $s_t$ and action $a_t$, the corresponding $\mathcal{Q}$- value $\mathcal{Q}(s_t, a_t)$ is updated using temporal differences (TD).
A temporal difference describes the prediction error of a $\mathcal{Q}$-function with respect to an observed true reward $r_t$ as $TD(s_t, a_t)=r_{t} + \gamma \max\mathcal{Q}(s_{t+1}, \cdot)-\mathcal{Q}(s_t, a_t)$, where $\gamma$ is the \emph{discounting factor} that determines how strongly to weigh future rewards in the prediction.
For example, with $\gamma=0$, the temporal difference would describe the error in predicting immediate rewards without the influence of potential future rewards.
The estimate of the $\mathcal{Q}$ value can then simply be updated using temporal differences as $\mathcal{Q}(s_t, a_t) \gets \mathcal{Q}(s_t, a_t) +  \alpha TD(s_t, a_t)$ with $\alpha$ giving the \emph{learning rate}.
A policy can then be defined by only using the learned $\mathcal{Q}$-function as $\pi(s) = \argmax_{a\in\mathcal{A}}\mathcal{Q}(s,\cdot)$.
To ensure that the state space is sufficiently explored during learning, it is common to employ $\epsilon$-greedy exploration, where with probability $\epsilon$ an action $a_t$ is replaced with a random choice.

\citet{mnih2013playing} introduced deep $\mathcal{Q}$-networks (DQN) which models the $\mathcal{Q}$-function using a neural network and demonstrated its effectiveness in learning $\mathcal{Q}$-functions for complex, high-dimensional state spaces, such as video game frames.
However, \citet{van2016deep} identified that using a single network for action selection and value prediction when computing the temporal difference often creates training instabilities due to value overestimation.
They proposed to address this issue by using two copies of the network weights: one for selecting the maximizing action and another for value prediction.
The second set of weights remains static for brief periods before being updated with the values of the first set.
This \emph{double deep} $\mathcal{Q}$-network (DDQN) typically reduces overestimation bias and thereby stabilizes learning.
This advantage has also helped establish DDQN as one of the most widely used solution approaches in DAC.

\begin{algorithm2e}[t]%
$x \gets$ a sample from $\{0,1\}^{n}$ chosen uniformly at random\;

\For{$t \in \N$}{
$s \gets$ current state of the algorithm\;\label{line:getState}
$\lambda =\pi(s)$\;
$p = \lfloor \lambda \rceil/n$; and $c = 1/\lfloor \lambda \rceil$\;
\underline{\textbf{Mutation phase:}}\\
	Sample $\ell$ from $\Bin_{>0}(n,p)$\;
	\lFor{$i=1, \ldots, \lfloor \lambda \rceil$}
         {$x^{(i)} \assign \flip_{\ell}(x)$; Evaluate $f(x^{(i)})$}
	Choose $x' \in \{x^{(1)}, \ldots, x^{(\lfloor \lambda \rceil)}\}$ with $f(x')=\max\{f(x^{(1)}), \ldots, f(x^{(\lfloor \lambda \rceil)})\}$ u.a.r.\;
\underline{\textbf{Crossover phase:}}\\
\For{$i=1, \ldots, \lfloor \lambda \rceil$}
{$y^{(i)} \assign \cross_{c}(x,x')$\; 
\lIf{$y^{(i)} \notin \{x,x'\}$}{evaluate $f(y^{(i)})$}}
Choose $y' \in \{y^{(1)}, \ldots, y^{(\lfloor \lambda \rceil)}\}$ with 
    $f(y') = \max\{f(y^{(1)}), \ldots, f(y^{(\lfloor \lambda \rceil)})\}$ u.a.r.\;
\underline{\textbf{Selection and update step:}}\\
\lIf{$f(y') > f(x')$}{$y \assign y'$ \textbf{ else } $y \assign x'$}
\lIf{$f(y)\ge f(x)$}{$x \assign y$} 
}
\caption{The \onell with state space~$\mathcal{S}$, discrete portfolio $\K \coloneq \{ 2^i \mid 2^i \leq n \wedge i \in [0..k-1] \}$, and parameter control policy $\pi\colon\mathcal{S} \to \K$, maximizing a function $f\colon \{0,1\}^n \to \R$. $\lfloor \lambda \rceil \coloneq \lfloor \lambda \rfloor \text{ if } \lambda - \lfloor \lambda \rfloor < 0.5, \text{ else } \lceil \lambda \rceil$.
}
\label{alg:onell}
\end{algorithm2e}

DACBench \cite{eimer-ijcai21} provides a standardized collection of DAC problems, including both artificial benchmarks abstracting algorithm runs and real-world benchmarks from various AI domains. The complexity of DAC problems makes establishing ground truth difficult beyond artificial cases, limiting DACBench's capability to evaluate learned policies on real algorithms. This limitation was highlighted when \citet{benjamins2024instance} found that cross-instance policies unexpectedly outperformed instance-specific policies designed as performance upper bounds, potentially due to local optima. This underscores the need for ground-truth benchmarks to better understand DAC solutions. While new artificial benchmarks continue emerging \cite[such as, ][]{bordne-automlws24a}, theory-inspired DAC benchmarks \cite{biedenkapp2022theory,chen2023using} offer a promising middle ground, using theoretical insights from parameter control to provide optimality ground truth while maintaining real algorithm runs. The \leadingones benchmark \cite{biedenkapp2022theory} demonstrated this utility by revealing DDQN-based approaches' effectiveness in learning optimal policies for small action spaces while showing limitations with increased dimensionality.

In~\cite{chen2023using}, the \onemaxdac benchmark was introduced. Here, the goal is to control the parameter $\lambda$ of the \onell (Algorithm~\ref{alg:onell}) optimizing instances of the \onemax problem $\{f_z:\{0,1\}^n \rightarrow \R, x \mapsto \sum_{i=1}^n{x_i=z_i}\}$, also known as 2-color Mastermind. The parameter $\lambda$ determines the population size of the mutation and crossover phase, the mutation rate $p$, and the crossover bias $c$. Optimally controlling $\lambda$ as a function of the current-best fitness is a well-studied problem in the theory community, for which a policy for configuring $\lambda \in \R$ resulting in asymptotically optimal linear expected optimization time was derived as $\pi_{\texttt{cont}}(x) \coloneq \sqrt{\frac{n}{(n-f(x))}}$~\cite{doerr2015black,doerr2018optimal}.
The study of \onemaxdac in~\cite{chen2023using} highlighted the difficulty of this benchmark. 
Although a tailor-made approach based on \irace was able to find policies that performed on par with theoretical policies, its blackbox and ``cascading'' nature makes it highly \emph{sample inefficient}.

\section{DDQN for Solving \onemaxdac}
\label{sec:ddqn_for_onemax}

Compared to black-box DAC approaches, such as those based on \irace, deep-RL algorithms, especially off-policy algorithms like DDQN, are expected to be much more sample-efficient, since the learned policies can be updated \emph{during} every episode. This property makes DDQN and similar approaches appealing for DAC scenarios where each solution evaluation is expensive. However, deep-RL is commonly known to be difficult to use \citep[see, e.g.,][]{parker-holder-jair22a}. In this section, we investigate the learning ability of DDQN on the \onemaxdac benchmark. We will show that a na\"ive application of this commonly used deep-RL algorithm with a straightforward reward function results in limited learning ability. This motivates our study on reward function design in the next sections. 

\textbf{Action Space.} As DDQN is designed to work with a discrete action space, we discretize the \onemaxdac benchmark action space. For a given problem size $n$, we define the set of possible $\lambda$ values that the RL agent can choose from as $\{ 2^i \mid 2^i \leq n, i \in [0..\left\lfloor \log_2 n \right\rfloor] \}$. With this new action space, we define the following discretized version $\pi_{\texttt{disc}}(x)$ of the theory-derived policy: for a given solution $x$, we choose the \lbd value from the set that is the closest to $\pi_{\texttt{cont}}(x)$. 
\begin{figure}[t]
\centering
\begin{subfigure}[t]{0.49\linewidth}
    \includegraphics[width=\textwidth]{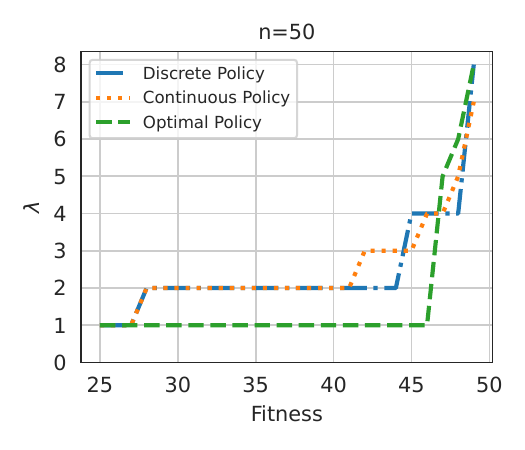}
\end{subfigure}
\begin{subfigure}[t]{0.49\linewidth}
    \includegraphics[width=\textwidth]{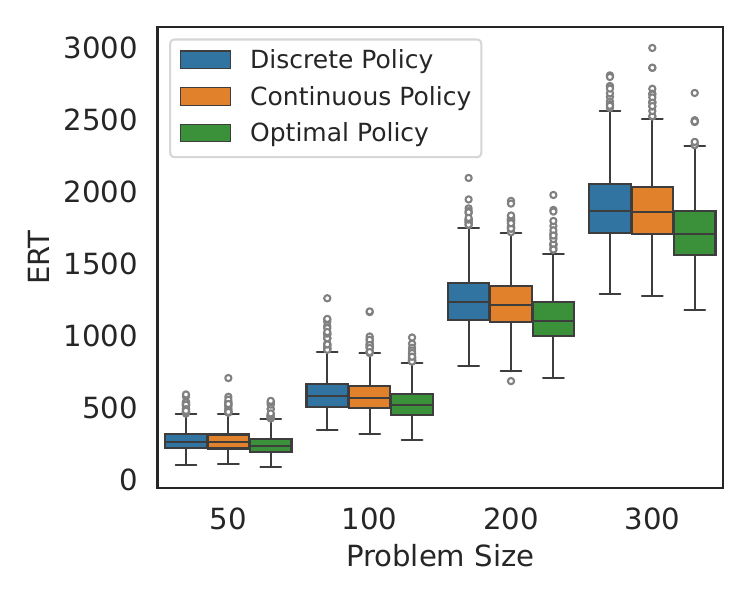}
\end{subfigure}
\caption{theory-derived policy, its discretized version, and the optimal policy for $f(x) \ge n/2$ with $n=50$ (left); their average runtimes across 4 problem sizes over $1{,}000$ runs (right).}
\label{fig:expected_runtime_theory}
\end{figure}

\Cref{fig:expected_runtime_theory} shows that the difference in performance between $\pi_{\texttt{cont}}$ and $\pi_{\texttt{disc}}$ is marginal (and not statistically significant according to a paired t-test with a confidence level of $99\%$). Therefore, an RL agent using this discretized action space should be able to find a policy that is at least competitive with the theory-derived one. Additionally, there is a gap between the optimal policy ($\pi_{\texttt{opt}}$) \cite{chen2023using} and $\pi_{\texttt{cont}}$ across all problem sizes. Therefore, we aim to propose an RL-based approach to produce a policy closer to $\pi_{\texttt{opt}}$.

\textbf{State Space.} Following both theoretical and empirical work on the benchmark~\cite{doerr2015optimal,chen2023using}, we only consider the state space defined by the quality (``fitness'') of the current-best solution; in the absence of ground-truth, more complex state spaces are left for future work.

\textbf{Reward Function.} The aim of the \onemaxdac benchmark is to find a policy that minimises the runtime of the \onell algorithm, i.e., the number of solution evaluations until the algorithm reaches the optimal solution. Therefore, an obvious component of the reward function is the number of solution evaluations at each time step (i.e., iteration) of the algorithm. 

Additionally, to reduce time collecting samples from bad policies during the learning, following ~\cite{biedenkapp2022theory}, we impose a cutoff time on each run of the \onell algorithm, which allows an episode to be terminated even before an optimal \onemax solution is reached. To distinguish the performance between runs that are terminated due to the cutoff time, we define the reward function as:
\begin{equation}
\label{eq:delta_minus_total_evals}
r_t = \Delta f_t -E_t
\end{equation}
where $E_t$ is the total number of solution evaluations at time step~$t$ and $\Delta f_t = f(x_t) - f(x_{t-1})$ represents the fitness improvement between the current and the previous time steps. 

\textbf{Baseline Policies.} We consider three baseline policies in our study, including the theory-derived policy $\pi_{\texttt{cont}}(x)$, its discretized version $\pi_{\texttt{disc}}(x)$, and the (near) optimal policy $\pi_{\texttt{opt}}(x)$ from~\cite{chen2023using}.

\textbf{Experimental Setup.} We train DDQN on four \onemax problem sizes $\{50, 100, 200, 300\}$. For each size, we repeat each RL training $10$ times using a budget of $500{,}000$ training steps. The training used a machine equipped with two dual-socket Intel\textsuperscript{\textregistered} Xeon\textsuperscript{\textregistered} E5-2695 v4 CPUs  (2.10 GHz), each with a maximum of 72 threads. We use a single thread for training and parallelize 20 threads for evaluation. A cutoff time of $0.8 n^2$, sufficiently larger than the optimal linear running time, is imposed on each episode during the training. 

Following ~\cite{biedenkapp2022theory}, we adopt a default hyperparameter configuration of DDQN with $\epsilon$-greedy exploration and $\epsilon=0.2$. The replay buffer size is set to 1 million transitions. At the beginning of the training process, we sample $10{,}000$ transitions uniformly at random and add them to the replay buffer before learning begins. The Adam optimiser \cite{adam} is used for the training process with a (mini-)batch size of $2{,}048$ and a learning rate of $0.001$. To update the $\mathcal{Q}$-network, we adopt a discount factor of $0.99$ and use the soft target update mechanism with $\tau = 0.01$ to synchronise the online policy and the target policy \cite{ddpg-soft-update}. 

\textbf{Performance Metrics.} To study the learning performance of each RL training, we record the learned policies at every $2{,}000$ training steps and evaluate each of them with $100$ different random seeds. We then measure the performance of each RL training via three complementary metrics:

\textbf{(1) Best policy's performance (ERT and gap):}  evaluate top $5$ policies across $1000$ random seeds, select the best performer, and compute its \textit{expected runtime (ERT)} and \textit{gap relative to baseline policy's ERT}.
    
\begin{figure}[t]
\centering
    \begin{subfigure}[t]{0.49\linewidth}
        \includegraphics[width=\textwidth]{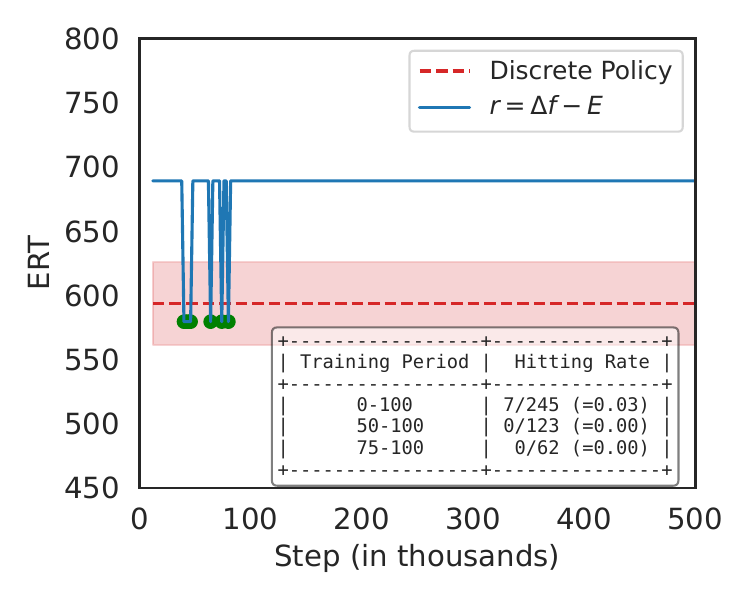}
    \end{subfigure}
    \begin{subfigure}[t]{0.49\linewidth}
        \centering
        \includegraphics[width=\linewidth, trim=0 0 0 0, clip]{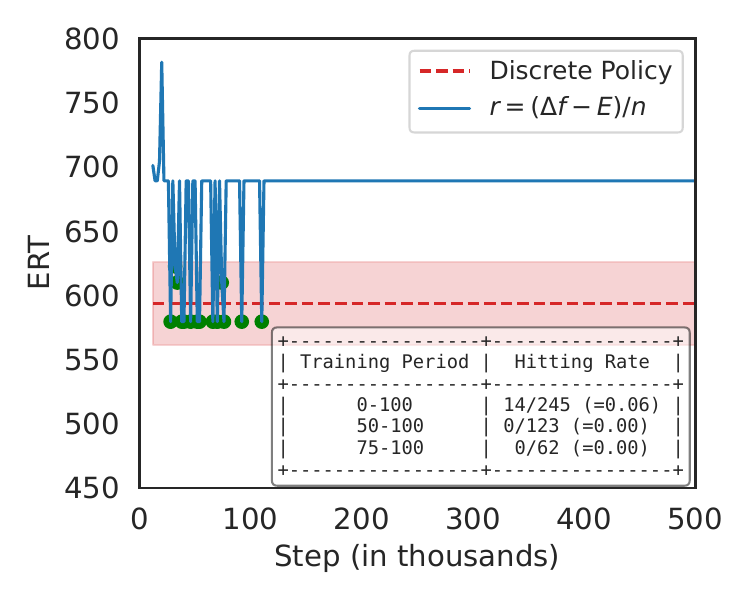}
    \end{subfigure}
    \begin{subfigure}[t]{0.49\linewidth}
        \centering
        \includegraphics[width=\linewidth, trim=0 0 0 0, clip]{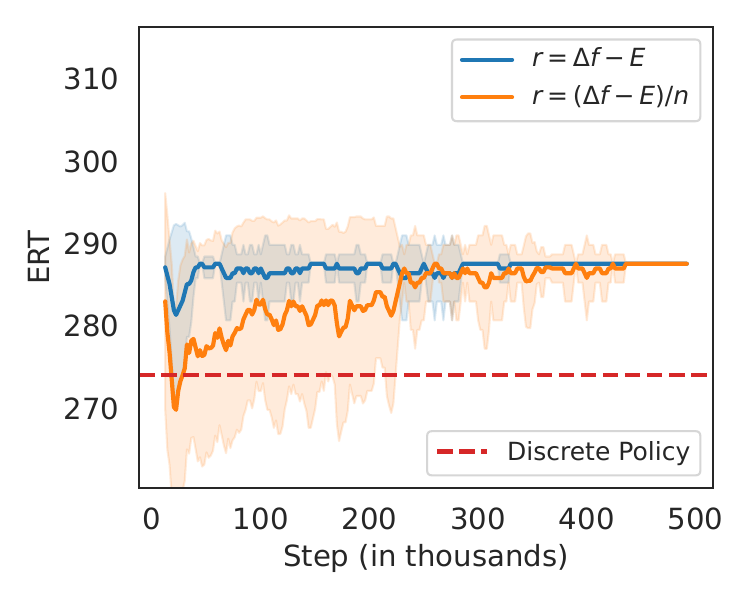}
    \end{subfigure}
    \begin{subfigure}[t]{0.49\linewidth}
        \centering
        \includegraphics[width=\linewidth, trim=0 0 0 0, clip]{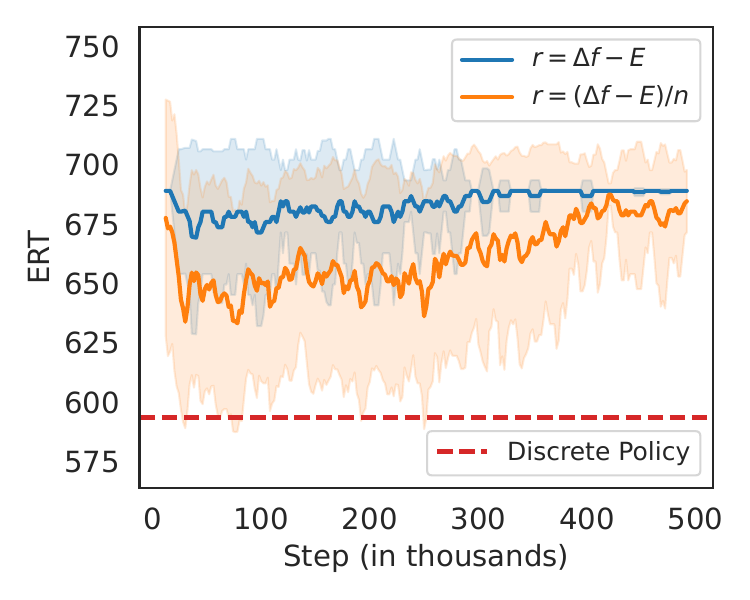}
    \end{subfigure}
    \caption{Top row: example DDQN learning curves on $n=100$ using original (left) and scaled (right) reward functions, with $\pi_{\texttt{disc}}$ as the baseline policy. Green dots indicate hitting points (details in ~\Cref{sec:ddqn_for_onemax}). Bottom row: average performance across 10 learning curves on $n=50$ (left) and $n=100$ (right).}
    \label{fig:learning_curve_naive_and_scale_n100}
\end{figure}

\textbf{(2) Area Under the Curve (AUC):} difference between the learning curve and the baseline performance of $\pi_{\texttt{disc}}(x)$ in ~\Cref{fig:learning_curve_naive_and_scale_n100} (top row), where the curve points represent the average runtime of the current policy across $100$ seeds. 

\textbf{(3) Hitting rate (HR):} ratio $n_h / n_e$ of policies with runtime within $\mu \pm 0.25\sigma$ of the baseline policy (where $\mu$, $\sigma$ are the ERT and standard deviation)~\cite{biedenkapp2022theory}, with hitting points $n_h$ shown as green dots in ~\Cref{fig:learning_curve_naive_and_scale_n100} and $n_e$ the total number of evaluations.
    
We aim to minimise ERT (gap) and AUC, while maximising HR.

\textbf{Na\"ive Rewards Fail to Scale.} \Cref{fig:gap_auc_n100_original_vs_scaled} (left plot) depicts the gap of the best-learned policy to $\pi_{\texttt{disc}}$ across $10$ RL runs using the reward function defined in ~\Cref{eq:delta_minus_total_evals}.
The blue box plots highlight the limitation in the DDQN learning scalability: Although the agent can find policies of reasonable quality in the smallest problem size of $n=50$, the gap increases significantly with $n$. From $n=200$ onward, the agent is no longer able to get close to the baseline during the whole learning process.

\textbf{Na\"ive Rewards Induce Learning Divergence.} 
\Cref{fig:learning_curve_naive_and_scale_n100} (top left) shows an example of a DDQN learning curve on $n=100$ where the agent is able to find a good policy at the beginning but then starts to diverge and stagnate until the end of the learning process. This behaviour is further demonstrated in the bottom row of ~\Cref{fig:learning_curve_naive_and_scale_n100}, where we show the agent's average ERT (blue lines). The agent consistently diverges in the later part of the training process across all RL runs on both problem sizes. This issue makes the RL agent incapable of exploiting knowledge of well-performing policies, likely leaving a user with a far-from-optimal policy at the end of the training process, especially when we have a limited budget and cannot afford thorough evaluations to select the best-performing policy.

\section{Reward Scaling}
\label{sec:reward_scaling}

\begin{figure}
    \centering
    \includegraphics[width=0.49\linewidth, trim=0 0 0 0, clip]{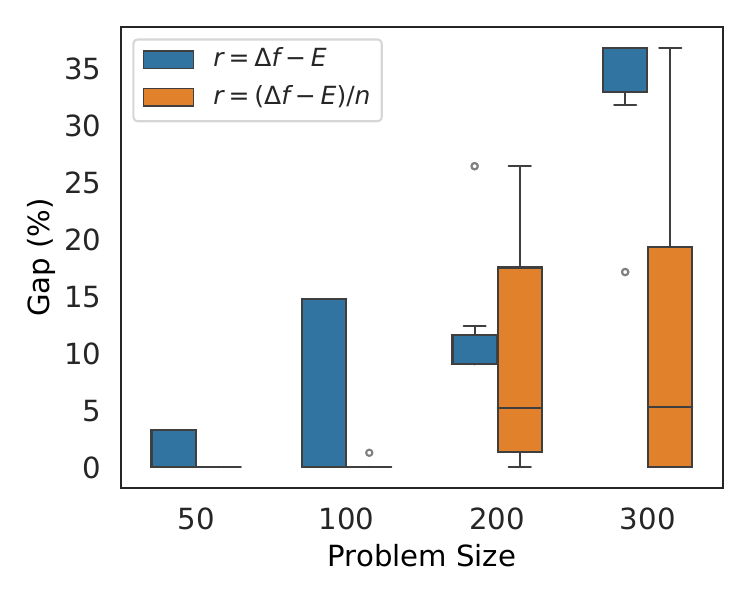}
    \includegraphics[width=0.49\linewidth, trim=0 0 0 0, clip]{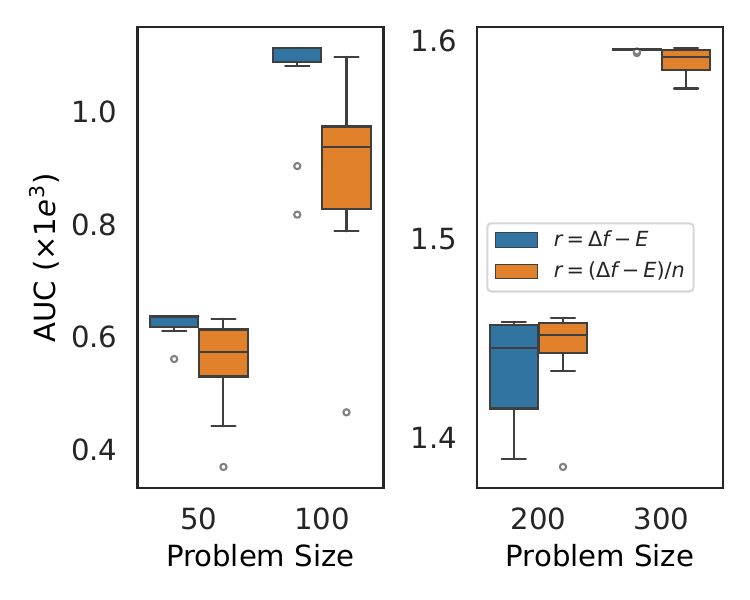}
    \caption{DDQN performance (as gap to $\pi_{\texttt{disc}}$ and AUC) using original and scaled reward functions.}
    \label{fig:gap_auc_n100_original_vs_scaled}
\end{figure}

\begin{figure}
    \centering
    \includegraphics[width=0.49\linewidth, trim=0 0 0 0, clip]{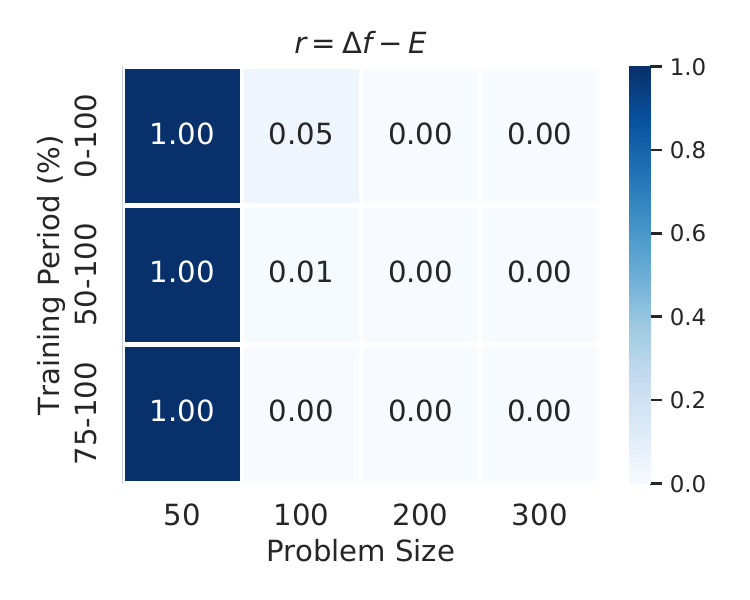}
    \includegraphics[width=0.49\linewidth, trim=0 0 0 0, clip]{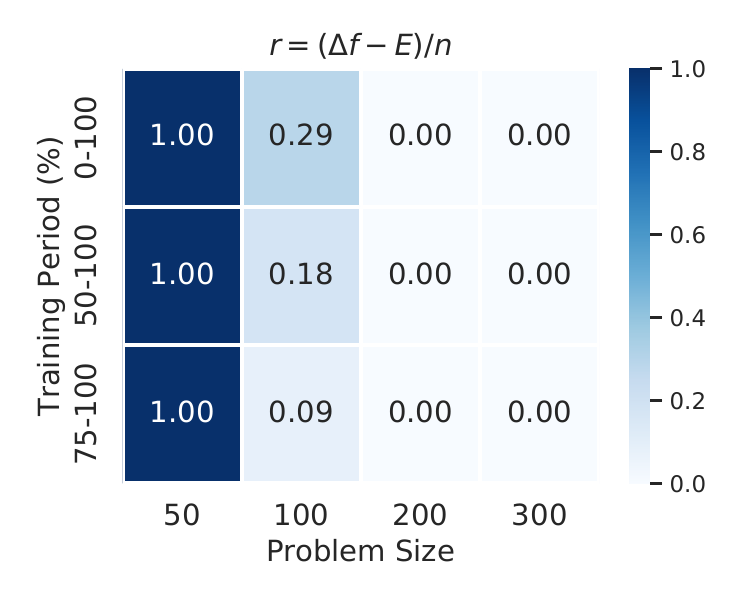}
    \caption{DDQN performance (as HR across different training periods) using original and scaled reward functions.}
    \label{fig:heatmap_n100_original_vs_scaled}
\end{figure}

In the previous section, we have shown two learning limitations of DDQN on the \onemaxdac benchmark. In this section, we study the impact of \emph{reward scaling} on DDQN learning performance. Our study is inspired by the fact that the original reward function $r_t = \Delta f_t - E_t$ is significantly influenced by the selected action. More concretely, from Algorithm~\ref{alg:onell}, we can infer that $E_t \approx 2\lambda$. To maximise the returns, the agent would be biased towards choosing actions that minimise $E_t$, resulting in a bias towards choosing smaller $\lambda$ values. 
In fact, we see that during the divergence period, the agent gets stuck consistently in policies that dominantly select $\lambda=1$ (the smallest $\lambda$ in our action space) across the entire state space. 
Moreover, the wide variance between the values of $\lambda$ in the portfolio poses a significant challenge for the agent to learn its behavior from its environment. There is a need to limit the reward range, where the normalization mechanism is proved as a simple yet effective solution. The normalization process enhances the $\mathcal{Q}$-network's ability to discover the appropriate parameters $\theta$ much more efficiently \cite{van2016learning,sullivan2024reward}. More specifically, to mitigate the variance scaling of $\lambda$ in the portfolio as the problem size increases, we scale the original reward function by the problem size:
\begin{equation}
\label{eq:imp_minus_evals_problem_scaled}
r_t = \Big( \Delta f_t - E_t \Big)/n
\end{equation}
The scaled reward value thus ranges from $\frac{-2 \text{max}(\lambda)}{n}$ to $\frac{(n-1)}{n}$.

\begin{table}[t]
    \centering
    \caption{Comparison of ERT (and its standard deviation) among theory-derived policies, RL-based DAC with two designed reward functions (top), and the optimal policy (bottom) across four problem dimensions. Bold values indicate the best ERT among theory-derived and RL-based policies.}
    \label{tab:comparison_2_rw_fncs}
    \begin{adjustbox}{max width=0.48\textwidth} 

    \begin{tabular}{l cccc}
        \toprule
         & \multicolumn{4}{c}{\textbf{ERT}($\downarrow$)} \\
        \cmidrule(lr){2-5}
        & $n=50$ & $n=100$ & $n=200$ & $n=300$ \\
        \midrule
        $\pi_{\texttt{cont}}$ & 272.39\scriptsize{(76.40)}& \textbf{582.62}\scriptsize{(118.45)}&\textbf{1233.38}\scriptsize{(193.01)} & \textbf{1883.42}\scriptsize{(252.22)} \\
        $\pi_{\texttt{disc}}$ & 274.02\scriptsize{(74.49)}&593.44\scriptsize{(128.13)}&1248.88\scriptsize{(195.01)} & 1889.45\scriptsize{(251.95)} \\
        $r = \Delta f - E$ & 271.38\scriptsize{(91.16)} & 643.67\scriptsize{(184.91)} & 1409.01\scriptsize{(305.43)} & 2500.03\scriptsize{(591.56)}\\
        $r = (\Delta f - E)/n$ & \textbf{255.24}\scriptsize{(78.06)} & 583.64\scriptsize{(139.03)} & 1364.49\scriptsize{(276.65)} &2077.52\scriptsize{(338.43)}\\
        \midrule
        $\pi_{\texttt{opt}}$& 246.39\scriptsize{(71.13)}& 531.27\scriptsize{(110.09)}& 1121.71\scriptsize{(177.79)} & 1725.01\scriptsize{(224.35)} \\
        \bottomrule
    \end{tabular}
    \end{adjustbox}

\end{table}

As shown in~\Cref{tab:comparison_2_rw_fncs}, the original reward function performs well in the small problem size of $50$. However, it becomes ineffective for the larger problem sizes. A similar observation is also made in the reward scaling, which are very promising for two smaller problem sizes, $n \in \{50,100\}$, but gradually becomes poor for larger problem sizes. \Cref{fig:gap_auc_n100_original_vs_scaled} shows that the scaling mechanism can help slightly in the cases of $n \in \{200,300\}$ by examining the gap percentage relative to the discrete theory. In particular, the reward scaling function performs exceptionally well compared to the conventional reward function. Looking at the problem size of $n=100$, which is highly competitive with the performance of both theory-derived policies. However, ~\Cref{fig:learning_curve_naive_and_scale_n100} reveals its ERT diverge significantly and the best expected runtimes reported in the ~\Cref{tab:comparison_2_rw_fncs} for the reward scaling come from the early phases of learning. 

We used a heat map to provide an overview of HR across $10$ RL runs. We analyze three distinct phases of the training process: the entire duration (0\%-100\%), the latter half (50\%-100\%), and the last quarter (75\%-100\%). By segmenting these periods, the heatmap helps identify potential divergence points. In an ideal scenario, where the RL agent progressively learns from the environment, we expect a consistent growth in the HR across all three phases. As shown in \Cref{fig:heatmap_n100_original_vs_scaled}, the smallest problem size achieves HR of 1 for all three periods, as the difference between good and poor policies in this setting is minimal. We thus need to analyze larger problems to gain a clearer understanding of the landscape. 

Generally, reward scaling effectively addresses scalability, particularly in the setting of problem size $n = 100$, where the agent discovers several good policies at the outset, achieving $\text{HR}@(0\%-100\%) = 0.29$ compared to $0.05$ with the original reward function. However, these HRs begin to decline over time, indicating the occurrence of divergence in both the original and scaled reward functions. More concretely, the heatmap for $n = 100$ under the scaled reward function reveals that HRs decrease over the three analyzed periods: $\text{HR}@(0\%-100\%) = 0.29$, $\text{HR}@(50\%-100\%) = 0.18$, and $\text{HR}@(75\%-100\%) = 0.09$. A similar observation is evident in the heatmap for the conventional reward signal. For problems of size $n = 200$ and $n = 300$, the divergence issue becomes more severe, as no green dots ($\text{HR} = 0$) are observed in all periods.

\section{An Under-exploration Issue}
\label{sec:under_exploration}

Learning with the original reward and the scaled reward functions is challenged by the \emph{divergence} problem. As shown in ~\Cref{fig:learning_curve_naive_and_scale_n100}, the RL agent tends to require more runtimes when more training is given, in contrast to our expectation. We conjecture that the agent lacks the ability to discover more rewarding actions. We call this issue \emph{divergence-stagnation}: the learned RL agent fails to find effective actions during the exploration phase, leading to the agent being trapped in a suboptimal policy or a collection of poor policies.

To understand how the learned RL agent gets trapped in poor policies, 
we examine how the policy changes during RL training for both original and scaled reward functions for $n=100$. We use a pairwise difference between two consecutive evaluated policies:
\begin{equation}
D(\pi_t, \pi_{t-1}) = \sum_{s \in \mathcal{S}} \mathbbm{1}[\pi_t(s) \neq \pi_{t-1}(s)] \quad
\end{equation}
where $\mathcal{S}$ is the full set of states in the environment, $t$ denotes the time step in the evaluation phase, and $\pi$ is the online policy.
A higher value of the difference between consecutive policies indicates the variety in RL training. We generally expect a large number of changes in the beginning, indicating that the agent is exploring. Later, the policy change is expected to decrease over time for learning convergence and stability. 

\begin{figure}[t]
    \centering
    \includegraphics[width=0.6\linewidth, trim=0 0 0 0, clip]{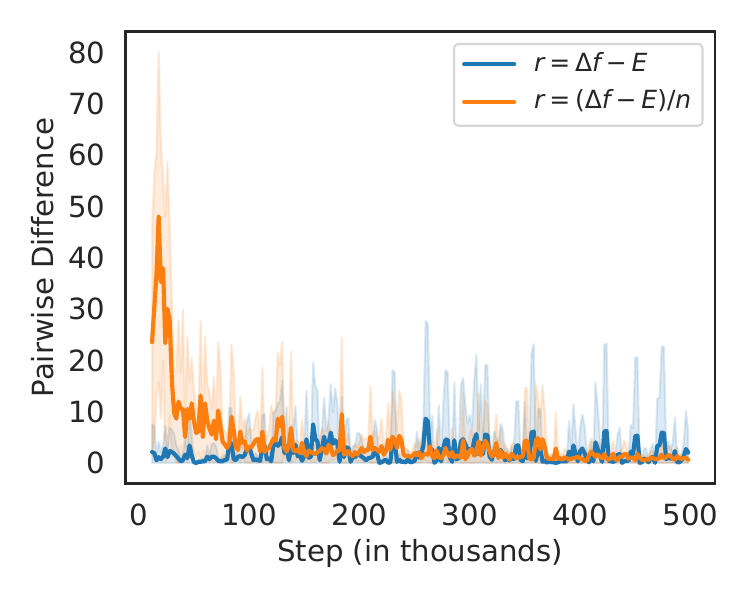}
    \caption{Example of the pairwise difference for two reward functions with problem size 100.}
    \label{fig:under_exploration}
\end{figure}

Learning with the original reward function (shown in ~\Cref{fig:under_exploration}) however lacks changes in the beginning: its pairwise difference is always stably close to zero. This explains why, in ~\Cref{fig:learning_curve_naive_and_scale_n100}, the agent struggles to decrease runtime and finally stagnates. Learning with the scaled reward function, on the other hand, follows our expected pattern, which is why it performs better in ~\Cref{fig:learning_curve_naive_and_scale_n100}. Nevertheless, the fact that it still can not overcome the divergence-stagnation challenge suggests that the agent does not explore the environment enough. We thus look for an effective solution to encourage exploration during learning.

\begin{figure}[t]

    \centering
    \begin{subfigure}[t]{0.49\linewidth}
        \centering
        \includegraphics[width=\linewidth, trim=0 0 0 0, clip]{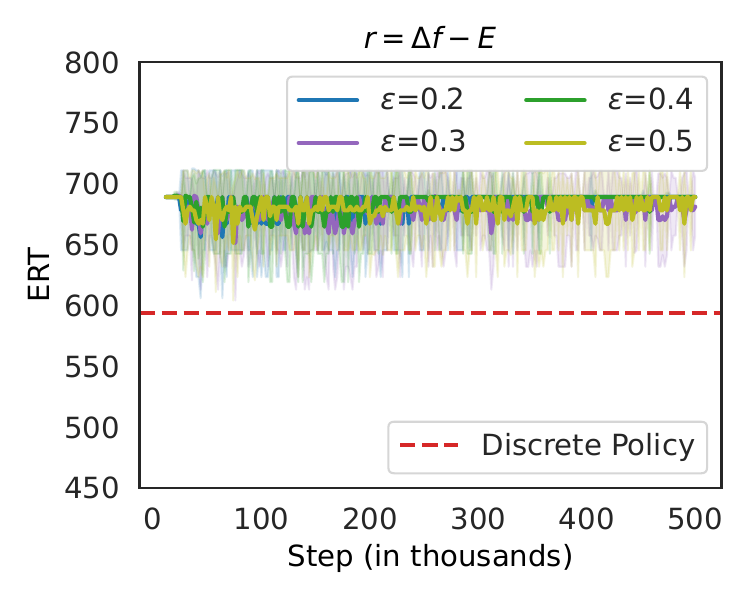}
    \end{subfigure}
    \begin{subfigure}[t]{0.49\linewidth}
        \centering
        \includegraphics[width=\linewidth, trim=0 0 0 0, clip]{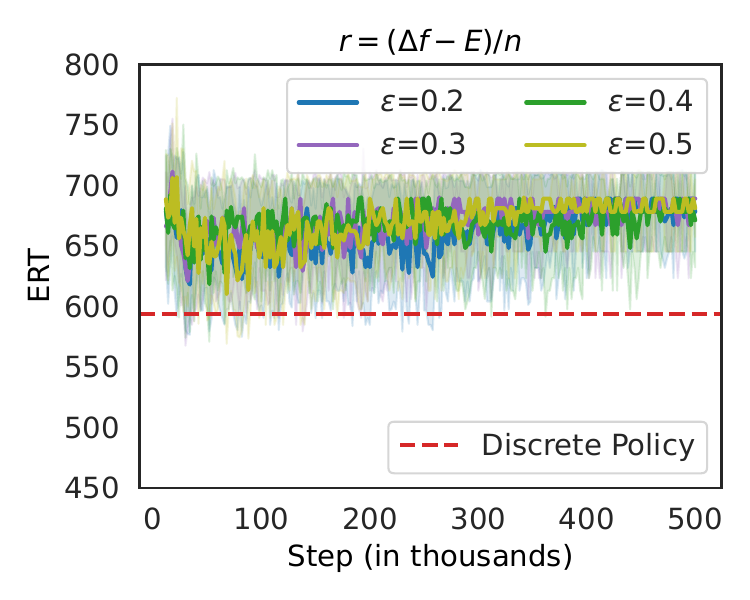}
    \end{subfigure}
    
    \centering
    \begin{subfigure}[t]{0.49\linewidth}
        \centering
        \includegraphics[width=\linewidth, trim=0 0 0 0, clip]{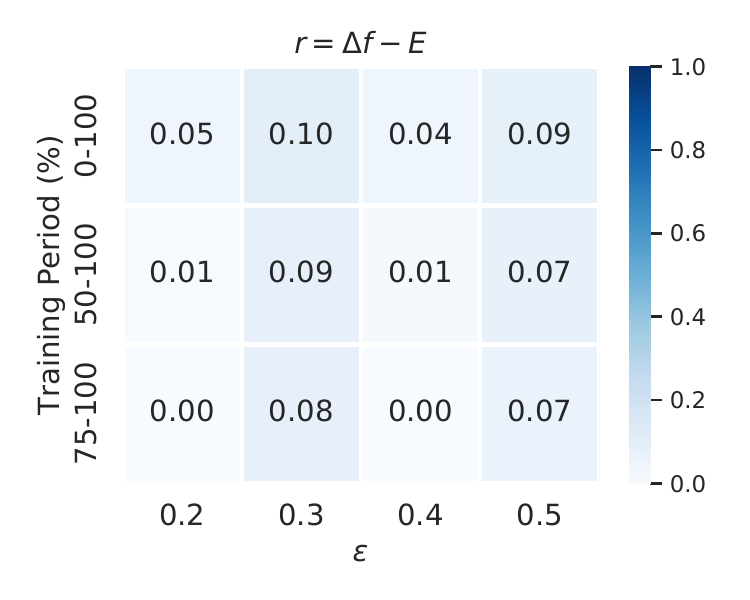}
    \end{subfigure}
    \begin{subfigure}[t]{0.49\linewidth}
        \centering
        \includegraphics[width=\linewidth, trim=0 0 0 0, clip]{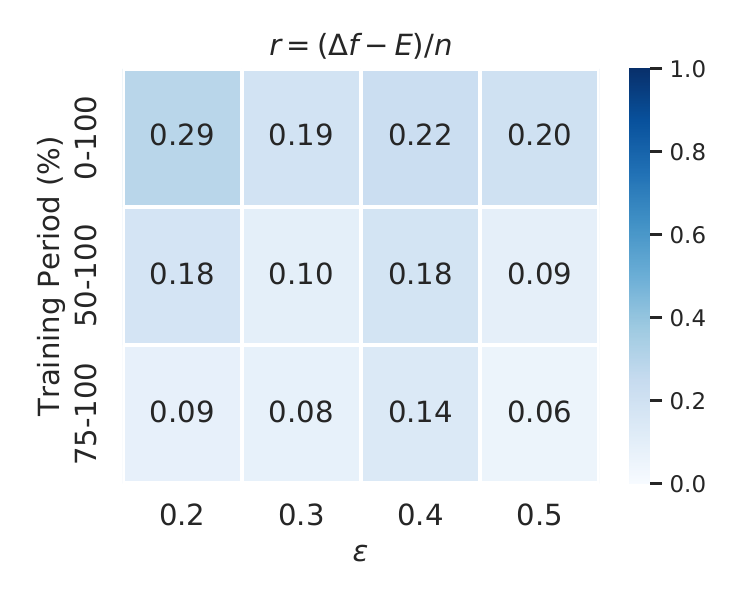}
    \end{subfigure}

    \begin{subfigure}[t]{0.49\linewidth}
        \centering
        \includegraphics[width=\linewidth, trim=0 0 0 0, clip]{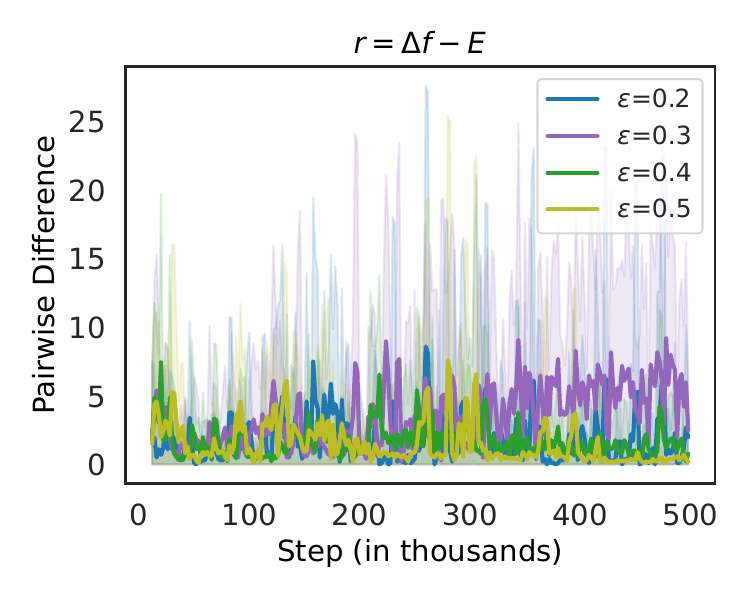}
    \end{subfigure}
    \begin{subfigure}[t]{0.49\linewidth}
        \centering
        \includegraphics[width=\linewidth, trim=0 0 0 0, clip]{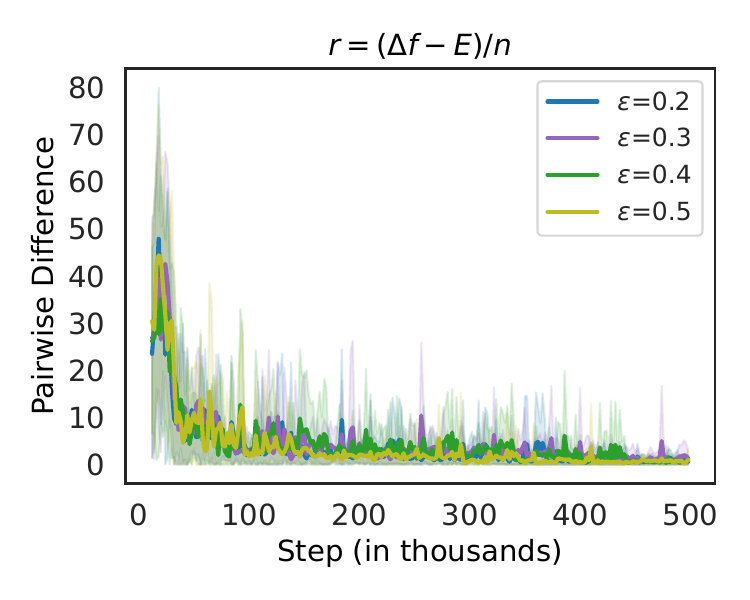}
    \end{subfigure}
    
    \caption{$\epsilon$-greedy with various $\epsilon$ values on $n=100$, applied on DDQN with original (left) and scaled (right) reward functions.
    }
    \label{fig:epsilon}
\end{figure}

\textbf{Random Exploration.} In the literature, a popular choice to encourage exploration in learning is $\epsilon$-greedy~\cite{sutton1988learning}. We tried various values of $\epsilon \in \{0.2, 0.3, 0.4, 0.5 \}$ for the problem $n=100$, using the DDQN setting described in~\Cref{sec:ddqn_for_onemax}. The top row in~\Cref{fig:epsilon} shows that the evaluated ERT still diverges until the end of the training budget, despite some slight improvements over the default setting ($\epsilon=0.2$). For instance, learning with the original reward function achieves the best HR when $\epsilon=0.3$, still the HR never exceeds $0.1$ (middle row in~\Cref{fig:epsilon}). 
We believe that the simple $\epsilon$-greedy strategy, which only explores the action space uniformly randomly, does not effectively improve exploration, as we can see in the bottom row in~\Cref{fig:epsilon} that the pairwise difference curves of different $\epsilon$ values are barely indistinguishable. This motivates our next study on employing a more sophisticated mechanism for improving exploration in deep-RL, namely \emph{reward shifting}~\cite{sun2022exploit}.
\section{Reward Shifting}
\label{sec:reward_shifting}

The life-long history of RL, combined with extensive experimentation in various environments, has revealed that relying solely on the $\epsilon$-greedy strategy is inadequate to address the issue of under-exploration~\cite{strehl2004empirical,bellemare2016unifying,osband2016deep,choshen2018dora}. These studies suggest more systematic approaches that encourage agents to prioritize visiting specific states, thereby improving exploration. Reward shaping~\cite{randlov1998learning,ng1999theory,laud2004theory} is a robust strategy in addressing the exploration challenge in RL training \cite{sun2022exploit,10.5555/3635637.3662910,dey2024continual,mareward}. The idea of reward shaping is to incorporate an external reward factor, $\mathcal{R}'=\mathcal{R}+F$ where $F$ is a shaping function, into the original reward $\mathcal{R}$. The external factor not only assists in stabilizing the RL training process but also accelerates the convergence of the RL algorithm. The stabilization arises from a sufficient trade-off between exploitation and exploration. Reward shaping can accelerate RL training because the agent needs to minimize the number of steps within an episode~\cite{burda2018large}. If the agent takes unnecessary steps, the cumulative reward becomes more negative, reducing the overall reward.

The effectiveness of reward shaping in addressing the hard exploration challenge lies in the fact that the $\mathcal{Q}$-network is trained under the assumption of \emph{optimistic initialization} \cite{brafman2002r,even2001convergence,szita2008many}. The values of the $\mathcal{Q}$-network are initialized with optimistically high values: $\tilde{\mathcal{Q}}_0(s, a; \theta) \geq \max_{a'} \mathcal{Q}^*(s, a') \; \forall (s, a)$, where $\tilde{\mathcal{Q}}_0(\cdot)$ represents the $\mathcal{Q}$-values of the approximate network at the initial step and $\mathcal{Q}^*(\cdot)$ is the true optimal $\mathcal{Q}$-values. During the training, the trained $\mathcal{Q}$-network gradually relaxes to estimates of the expected returns based on the observations.

Inspired by the theory of reward shaping, the work \cite{sun2022exploit} introduces \emph{reward shifting}: $\mathcal{R}'=\mathcal{R}+b$ with $b \in \mathbb{R}$, to encourage the agent to explore its environment, which helps the agent escape from a suboptimal point in the optimization landscape. In the context of reward shifting, it is not necessary to directly initialize the network's parameters to obtain higher $\mathcal{Q}$-values. Instead, the $\mathcal{Q}$-values of the optimal policy $\pi^{*}$ are shifted downward by a constant relative to the original situation. A proof of the relationship between external bias and the shifting distance is provided in the appendix. This shift assumes that all values associated with the approximate policy $\tilde{\pi}$ are initially higher than those of the true optimal policy $\pi^{*}$. In the network update, the $\mathcal{Q}$-value of the chosen action at step $t$ is pulled closer to the shifted true optimal policy. Meanwhile, the $\mathcal{Q}$-values of the unchosen actions are maintained at relatively high levels. In this manner, the selected action at step $t+1$ will less frequently adhere to the knowledge of the optimal policy, resulting in a more extensive exploration of the environment. In contrast, initializing with lower $\mathcal{Q}$-values of $\tilde{\pi}$ or setting the optimal policy $\pi^{*}$ higher can steer the agent toward exploitation rather than exploration; this concept is commonly referred to as \emph{pessimistic initialization}. By demonstrating some empirical results,  \citet{sun2022exploit} conclude that an upward shift is associated with a positive value of $b^{+}$, which leads to conservative exploitation, while a downward shift corresponds to a negative value of $b^{-}$, inducing curiosity-driven exploration.

\subsection{Reward Shifting with a Fixed Bias}

Following~\cite{sun2022exploit}, we implement the shifting mechanism by adding a constant bias into the original reward function in~\Cref{eq:delta_minus_total_evals}:
\begin{equation}
\label{eq:delta_minus_total_evals_bias}
r_t = \Delta f_t-E_t+b
\end{equation}
To determine the optimal shifting bias $b$, we replicate the experiments conducted in previous sections, employing both negative and positive biases. The considered shifts are $\{\pm 1, \pm 3, \pm 5\}$.

\begin{table}[t]
    \centering
    \caption{ERT of optimal, theory-derived and RL-based DAC policies with fixed and adaptive shifting $b_{a}^{-}$. Standard deviations are in parentheses. Blue: RL outperforms theory-based policies. Bold: best non-optimal ERT.}
    \label{tab:comparison_fix_reward_shifting}
    \begin{adjustbox}{max width=0.48\textwidth} 

    \begin{tabular}{l cccc}
        \toprule
         & \multicolumn{4}{c}{\textbf{ERT}($\downarrow$)} \\
        \cmidrule(lr){2-5}
        & $n=50$ & $n=100$ & $n=200$ & $n=300$ \\
        \midrule
        $\pi_{\texttt{cont}}$ & 272.39\scriptsize{(76.40)}&582.62\scriptsize{(118.45)}&1233.38\scriptsize{(193.01)} & 1883.42\scriptsize{(252.22)} \\
        $\pi_{\texttt{disc}}$ & 274.02\scriptsize{(74.49)}&593.44\scriptsize{(128.13)}&1248.88\scriptsize{(195.01)} & 1889.45\scriptsize{(251.95)} \\
        $r=\Delta f - E + 1$ & 282.88\scriptsize{(100.56)} & 680.96\scriptsize{(213.77)} & 1578.61\scriptsize{(408.37)} & 2580.92\scriptsize{(647.50)}\\
        $r=\Delta f - E + 3$ & 282.88\scriptsize{(100.56)} & 680.96\scriptsize{(213.77)} & 1578.61\scriptsize{(408.37)} & 2583.82\scriptsize{(648.63)}\\
        $r=\Delta f - E + 5$ & 282.43\scriptsize{(99.66)} & 678.59\scriptsize{(211.29)} & 1578.25\scriptsize{(409.83)} & 2563.81\scriptsize{(642.85)}\\

        $r=\Delta f - E - 1$ & \cellcolor{cyan!10} 249.68\scriptsize{(71.57)} &  \cellcolor{cyan!10}568.14\scriptsize{(132.78)} & 1315.25\scriptsize{(253.09)} & 2112.46\scriptsize{(368.38)}\\
        $r=\Delta f - E - 3$ &  \cellcolor{cyan!10}252.46\scriptsize{(70.54)} &  \cellcolor{cyan!10}542.11\scriptsize{(115.22)} &  \cellcolor{cyan!10}1195.08\scriptsize{(202.65)} & 1944.17\scriptsize{(302.85)}\\
        $r=\Delta f - E - 5$ &  \cellcolor{cyan!10}250.12\scriptsize{(70.92)} &  \cellcolor{cyan!10}551.99\scriptsize{(116.40)} &  \cellcolor{cyan!10}\textbf{1134.14}\scriptsize{(183.21)} & \cellcolor{cyan!10} 1835.33\scriptsize{(265.37)}\\

        $r=\Delta f - E + b^{-}_{\texttt{a}}$ & \cellcolor{cyan!10} \textbf{249.53}\scriptsize{(70.08)} & \cellcolor{cyan!10} 542.12\scriptsize{(120.62)} & \cellcolor{cyan!10} 1178.75\scriptsize{(200.03)} & \cellcolor{cyan!10} \textbf{1829.65}\scriptsize{(263.12)}\\
        $r=(\Delta f - E)/n + b^{-}_{\texttt{a}}$ & \cellcolor{cyan!10} 249.60\scriptsize{(71.23)} & \cellcolor{cyan!10} \textbf{538.25}\scriptsize{(122.48)} & \cellcolor{cyan!10} 1188.64\scriptsize{(209.49)} & \cellcolor{cyan!10} 1865.88\scriptsize{(293.33)}\\

        \midrule
        $\pi_{\texttt{opt}}$ & 246.39\scriptsize{(71.13)}& 531.27\scriptsize{(110.09)}& 1121.71\scriptsize{(177.79)} & 1725.01\scriptsize{(224.35)} \\
        \bottomrule
    \end{tabular}
    \end{adjustbox}

\end{table}

\textbf{Analysis.} In addition to using the ERT and HR to validate the effectiveness of incorporating the shifting bias into the original reward function of DDQN as in previous experiments, we also capture policy changes and the uncertainty in estimating action values to investigate the impact of the shifting mechanism on the diversification of policies. The uncertainty level is measured using the concept of entropy $\mathcal{H}(\pi) = - \sum_{a} \pi(a|s) \log \pi(a|s)$, where $\pi(a|s)$ represents the distribution of $\mathcal{Q}$-values across actions derived as $\pi(a|s) = \frac{\exp(\mathcal{Q}(s, a))}{\sum_{a'} \exp(\mathcal{Q}(s, a'))}$. A lower entropy indicates strong confidence in action-value estimation, while higher entropy suggests greater uncertainty, potentially encouraging more exploration.

\Cref{tab:comparison_fix_reward_shifting} shows the average ERT across $10$ RL runs for each approach, together with the baselines, where adding the positive fixed shifts $b$, which range from $+1$ to $+5$ results in generally higher ERT (i.e., worse performance). All settings associated with the negative biases outperform the positive options. More explicitly, they are better than the discretized theory policy $\pi_{\texttt{disc}}$ and even outperform the original theory policy $\pi_{\texttt{cont}}$ in several cases. This observation consolidates our conjecture about the under-exploration problem in using the original reward function, and that the RL agent should focus more on exploration than exploitation.

\begin{figure}[t]
    \centering

    \begin{subfigure}[t]{0.49\linewidth}
        \centering
        \includegraphics[width=\linewidth, trim=0 0 0 0, clip]{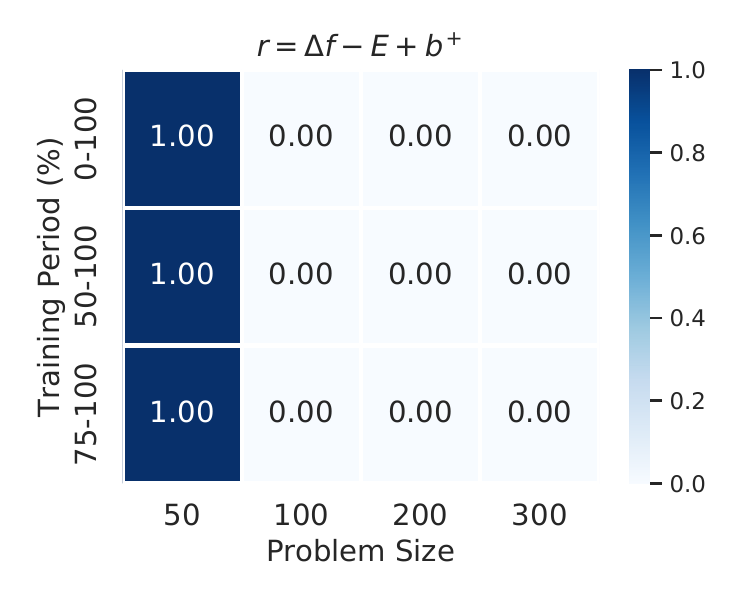}
    \end{subfigure}
    \begin{subfigure}[t]{0.49\linewidth}
        \centering
        \includegraphics[width=\linewidth, trim=0 0 0 0, clip]{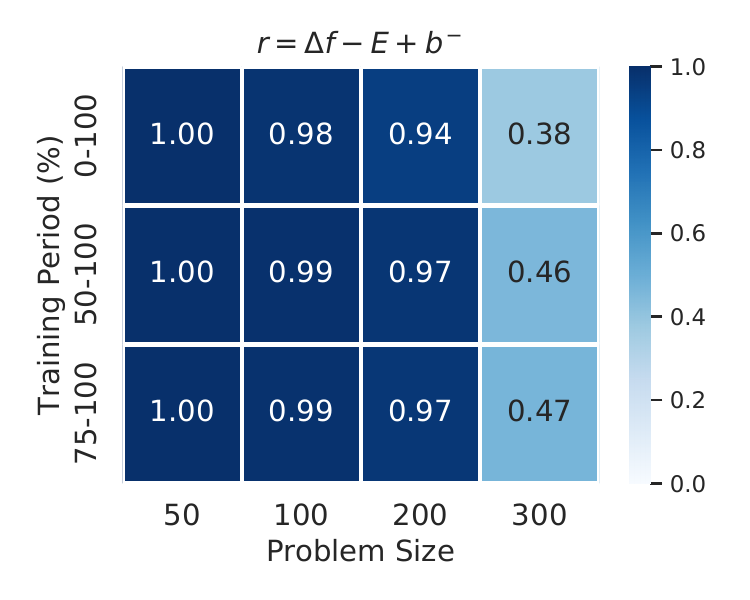}
    \end{subfigure}
    
    \caption{DDQN performance (HR across training periods) using the original reward function with fixed positive (left) and negative (right) shifts. For negative shifting, HRs of the best-performing biases for each problem size are presented.}
    \label{fig:gap_fixed_shifting}
\end{figure}

\Cref{fig:gap_fixed_shifting} provides more details about the ability of each approach in terms of converging to a good policy, where positive values of the shift completely fail the task with a problem size larger than $50$. In contrast, the robustness of the negative shiftings is demonstrated in the three problem sizes ranging from $50$ to $200$, where almost over $90\%$ of the evaluated points adhere to the theory policy. Although this strength is not maintained when the problem size increases to $300$, where the HR decreases by half, the negative reward shifting mechanism remains a promising solution to the stagnation problem as the HR does increase during the later part of the training process.

\begin{figure}[t]
    \centering
    \begin{subfigure}[t]{0.49\linewidth}
        \centering
        \includegraphics[width=\linewidth, trim=0 0 0 0, clip]{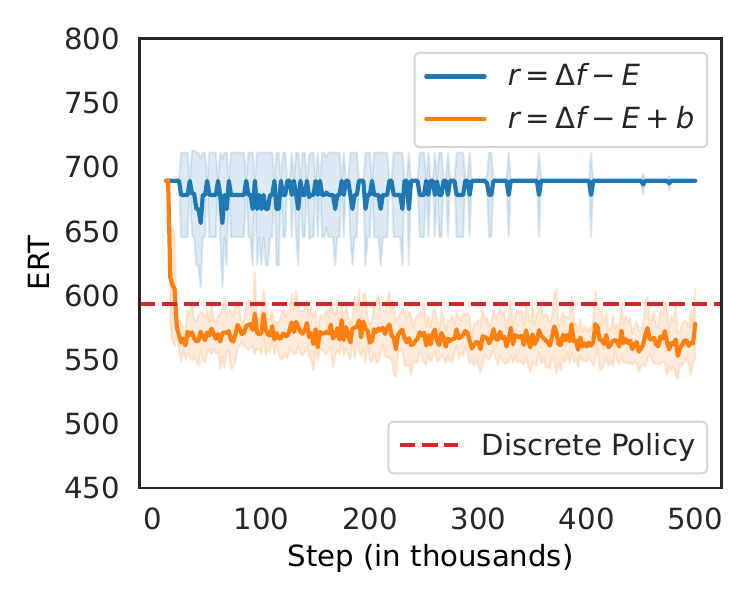}
    \end{subfigure}
    \begin{subfigure}[t]{0.49\linewidth}
        \centering
        \includegraphics[width=\linewidth, trim=0 0 0 0, clip]{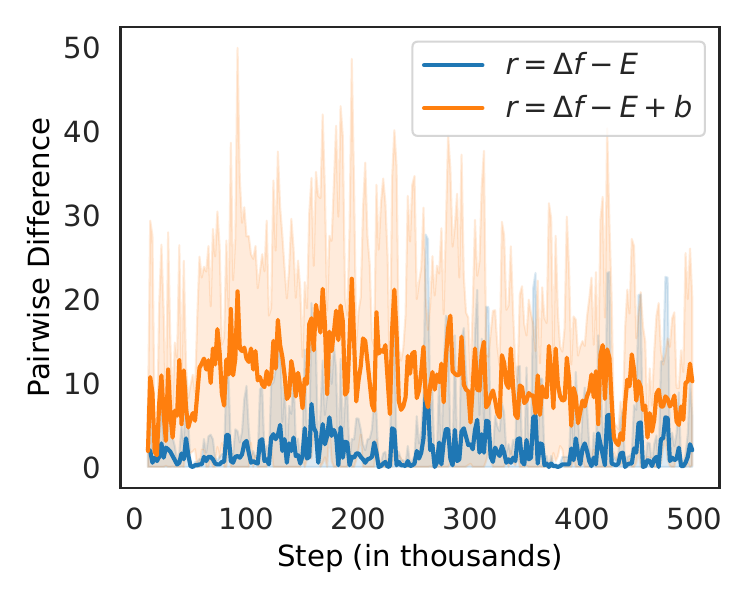}
    \end{subfigure}

    \begin{subfigure}[t]{0.49\linewidth}
        \centering
        \includegraphics[width=\linewidth, trim=0 0 0 0, clip]{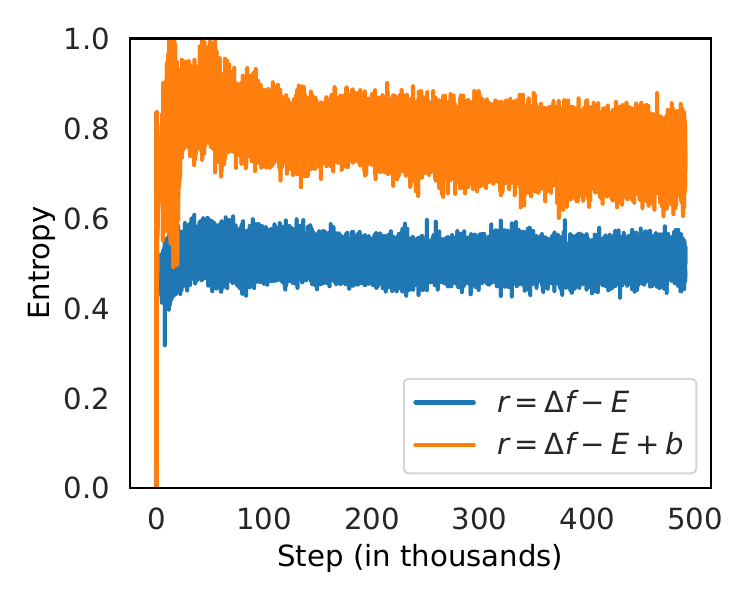}
    \end{subfigure}
    \begin{subfigure}[t]{0.49\linewidth}
        \centering
        \includegraphics[width=\linewidth, trim=0 0 0 0, clip]{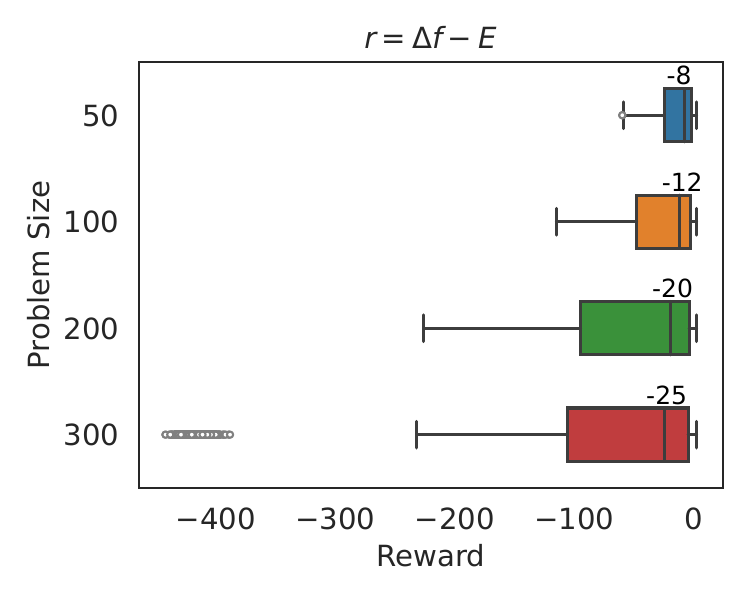}
    \end{subfigure}
    
    \caption{Effectiveness of reward shifting for problem size $100$ with $b = -3$. Top row: ERT and pairwise difference. Bottom row: entropy comparison and reward distribution for the first $10{,}000$ steps in a single DDQN run.}
    \label{fig:effectiveness_n100}
\end{figure}

In order to assess the effectiveness of reward shifting in helping the agent to explore its environment, we analyze the policy changes when employing the fixed shifting bias compared to the original reward function. ~\Cref{fig:effectiveness_n100} clearly demonstrates the significant diversity of policy changes when compared to those implemented without introducing shifting bias. In other words, the learned agent selects a variety of actions and gradually converges towards the desired outcome throughout the training process. Meanwhile, the agent walks in the correct direction toward the theory baseline as shown in the curve of ERT in~\Cref{fig:effectiveness_n100}. Furthermore, the entropy in action selection during the training process of adding reward bias is significantly higher than the entropy obtained from the original reward function. This observation suggests that the agent makes decisions with high uncertainty because it is aware of multiple paths that can lead to the goal. 

\subsection{Adaptive Shifting}
Choosing the right value for $b$ is a nontrivial task. As illustrated in~\Cref{tab:comparison_fix_reward_shifting}, the best bias varies across problem sizes. \citet{sun2022exploit} suggests using a \emph{meta-learner} to control the value of $b$, but this would significantly increase the cost of learning. Instead, we propose a simple yet effective mechanism, namely \emph{adaptive reward shifting}, which approximates the bias by leveraging the reward values obtained during the learning's warming-up phase, i.e., when DDQN collects random samples to initialize the replay buffer before the learning starts. 
\Cref{fig:effectiveness_n100} (bottom-right) shows the reward distribution in the replay buffer during this warmup phase across different problem sizes, where the number above each boxplot represents the median value. We define the adaptive bias $b^{-}_\texttt{a}$ as:
\begin{equation}
b^{-}_\texttt{a} = -0.2 |m(r)| \quad,
\end{equation}
where $m(r)$ is the median of the reward values observed during the warmup phase. The factor $0.2$ is decided based on comparing the medians in~\Cref{fig:effectiveness_n100} (bottom-right) and the best shifting values indicated in~\Cref{tab:comparison_fix_reward_shifting}. In practice, we may need to tune this factor when applying it to a new benchmark. However, this factor can be tuned across problem sizes or instances, in contrast to the fixed shifting value that likely needs to be tuned per problem size.

As shown in~\Cref{tab:comparison_fix_reward_shifting}, the proposed adaptive bias shifting, represented as $r_t = \Delta f_t - E_t + b^{-}_{\texttt{a}}$, achieves highly competitive ERT performance across all problem sizes. In the two cases of $n=50$ and $n=300$, the average ERT is even slightly better than the best ones obtained from fixed shifting values. 

\subsection{Reward Shifting and Scaling}

\begin{figure}[t]
    \centering
    \begin{subfigure}[t]{\linewidth}
        \centering
        \includegraphics[width=0.8\linewidth, trim=0 0 0 0, clip]{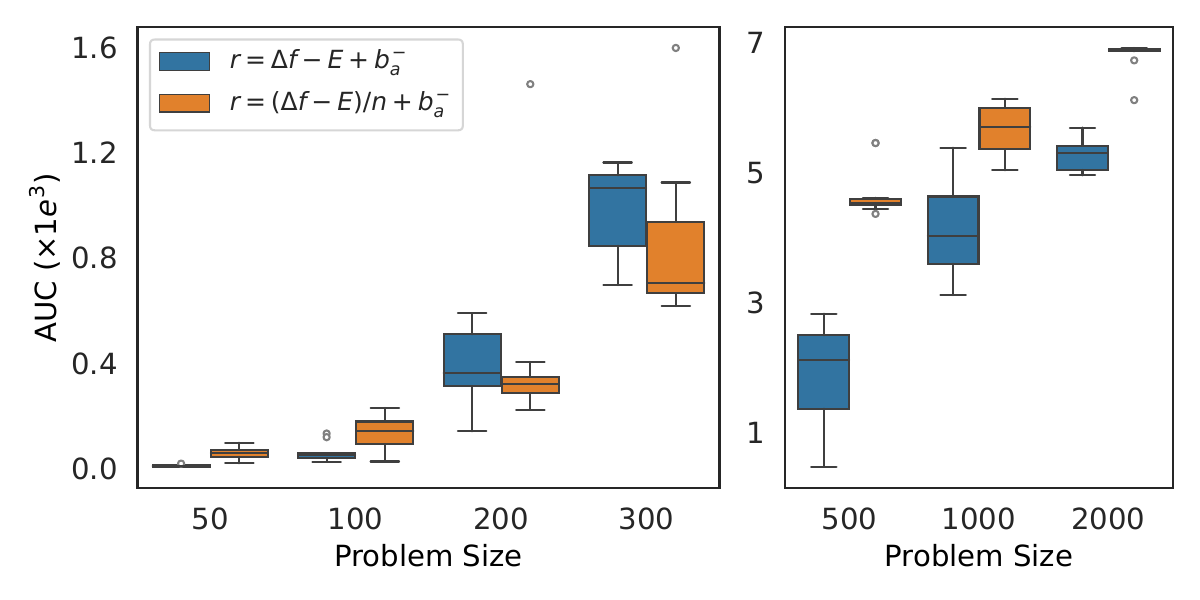}
    \end{subfigure}
    
    \begin{subfigure}[t]{\linewidth}
        \centering
        \includegraphics[width=0.8\linewidth, trim=0 0 0 0, clip]{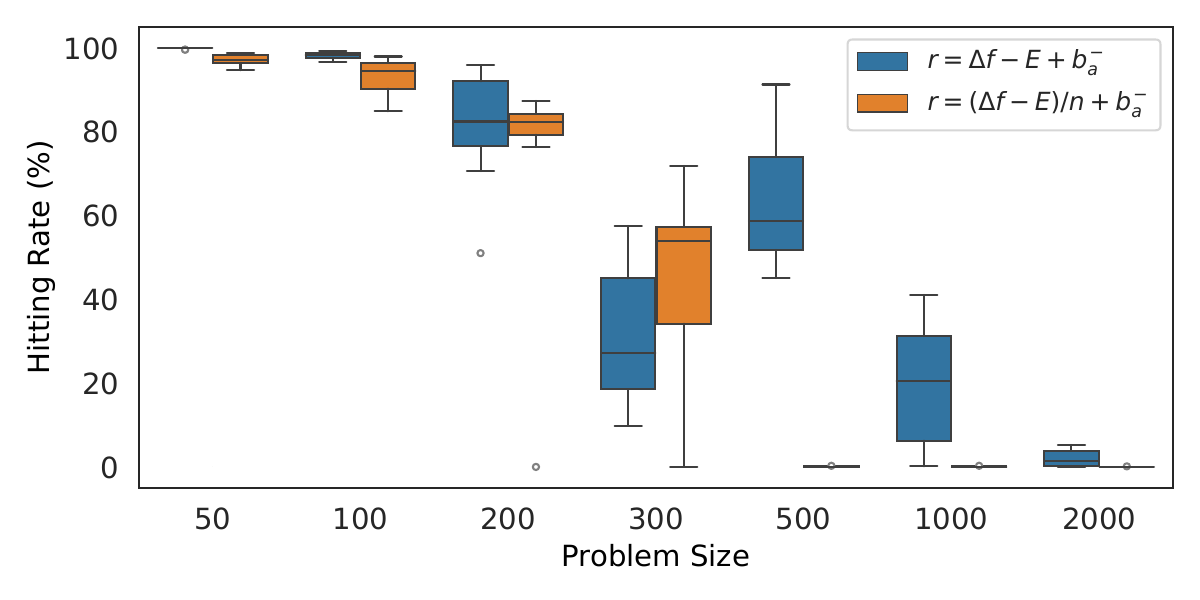}
    \end{subfigure}
    \caption{DDQN performance of adaptive shifting with the original and scaled reward functions across 7 problem sizes. 
    }
    \label{fig:compare_adaptive_shifting}
\end{figure}

In~\Cref{sec:reward_scaling}, we have observed some potential of the reward scaling technique to improve learning performance. It motivates our study in this section where we investigate the combined effect of reward shifting and reward scaling. As shown in the previous section, our proposed adaptive bias shifting demonstrates competitive performance compared to the best version of its fixed bias counterpart. We incorporate it into reward scaling: $r_t = \Big( \Delta f_t - E_t \Big)/n + b^{-}_{\texttt{a}}$.

As shown in~\Cref{tab:comparison_fix_reward_shifting}, compared to the shifting-only reward function, the overall average ERT of the shifting-scaling version is marginally worse. A similar observation concerning the learning stability, \Cref{fig:compare_adaptive_shifting} demonstrates that adding bias into the original reward surpasses the shifting-scaling approach in problem sizes of $n \in \{50, 100 \}$. In a problem size of $n=200$, the AUC and HR of the two reward functions are quite competitive with each other. On the other hand, the shifting-scaling association outperforms the original reward signal in the problem size of $n=300$. In this scenario, shifting-scaling achieves a higher HR compared to the conventional reward function. In general, we do not observe a clear advantage in combining the two mechanisms. 

\section{Scalability Analysis}
\label{sec:scalability_analysis}
We extend our experiments to larger problem sizes $n\in\{500,1000,2000\}$. In the previous section, we have seen that the magnitude of the bias should increase as the problem size increases. 
Instead of having to tune the bias for each new problem size, we adopt the adaptive shifting bias idea proposed in the previous section. 
Additionally, since the problem sizes are significantly larger, we increase the RL training budget to $1.5$ million steps, while keeping the same DDQN settings as in previous experiments.

\begin{table}[t]
    \centering
    \caption{Comparison of DDQN and the \irace cascading method~\cite{chen2023using} in terms of: 1) ERT and 2) the minimum number of time steps required for the found policy to surpass $\pi_{\texttt{disc}}$. 
    }
    \label{tab:comparison_n500_1000_2000}
    \begin{adjustbox}{max width=0.48\textwidth} 

    \begin{tabular}{l ccccccccc}
        \toprule
        & \multicolumn{2}{c}{$n=500$} & \multicolumn{2}{c}{$n=1000$} & \multicolumn{2}{c}{$n=2000$} \\
        \cmidrule(lr){2-7}
        & ERT & \#Steps & ERT & \#Steps & ERT & \#Steps \\
        &  & ($\times10^{6}$) &  & ($\times10^{6}$) &  & ($\times10^{6}$) \\
        \midrule
        $\pi_{\texttt{cont}}$ & 3237.16 & - & 6587.45 & - & 13361.92 & -\\
        $\pi_{\texttt{disc}}$ & 3271.48 & - & 6701.00 & -& 13642.12 & -\\

       $r=\Delta f - E + b^{-}_{\texttt{a}}$ & \textbf{3040.20} & \textbf{0.012}  & 6458.62 & \textbf{0.012} & 13599.94 & \textbf{0.012} \\
          $r=(\Delta f - E)/n+ b^{-}_{\texttt{a}}$ & 3592.75 & $\infty $ & 7046.52 & $\infty $ & 15855.36 & $\infty $ \\
        \irace & 3130.37 & $2002$ & \textbf{6421.68} & $20046$ & \textbf{12751.15} & $159439$\\
         \midrule
        $\pi_{\texttt{opt}}$ & 2938.08 & - & 6017.40 & - & 12185.07 & - \\
        \bottomrule

    \end{tabular}
    \end{adjustbox}

\end{table}

In~\Cref{tab:comparison_n500_1000_2000}, we present two metrics. First the average ERT of DDQN across 10 RL runs, compared to the best policies found using the \irace cascading method proposed in~\cite{chen2023using}. We observe that combining reward shifting and reward scaling provides no advantage over using reward shifting alone. This finding is further supported by~\Cref{fig:compare_adaptive_shifting}, where the AUC and HR values for the combination are generally lower than those for reward shifting alone.

Overall, our adaptive reward shifting mechanism achieves highly competitive ERT and consistently outperforms the discretized theory-derived baseline $\pi_{\texttt{disc}}$ across all three problem sizes. It also surpasses \irace on $n=500$ but performs worse than \irace on the larger problem sizes ($n=1000$ and $n=2000$). However, it is important to note that \irace's tuning budget is substantially larger than our RL training budget. For instance, on $n=500$, each iteration of \irace's cascading process consumes $5{,}000$ episodes, amounting to at least $5{,}000 \times 400 = 2{,}000{,}000$ time steps. There are 9 iterations in total, which results in an 18-million time-step tuning budget. The tuning budget increases to 75 millions and 308 millions for $n=1000$ and $n=2000$, respectively (compared to the 1.5 million budget of RL). The ERT reported in~\Cref{tab:comparison_n500_1000_2000} reflects the best result \irace found at the end of this extensive tuning process.

To better assess the sample efficiency of each approach, we measure the number of training time steps required to first surpass the discretized theory-derived baseline $\pi_{\texttt{disc}}$. This metric, shown in the \texttt{\#Steps} column of~\Cref{tab:comparison_n500_1000_2000}, reveals that DDQN with reward shifting requires several orders of magnitude fewer time steps than \irace across all three problem sizes, highlighting DDQN's strong advantage in sample efficiency.

\section{Conclusion}
\label{sec:conclusion}

Our study of deep reinforcement learning for dynamic configuration of \onell on \onemax reveals that na\"ive DDQN implementations face scalability and convergence issues due to the reward function design. We address these through reward scaling and shifting strategies, including automatic shifting bias adjustment. The proposed reward shifting approach achieves superior performance and significantly improved sample efficiency compared to the previous \irace-based method.

Although our results demonstrate the importance of reward design in DAC applications, future work could explore alternative reward shaping strategies and extend beyond the theoretical \onell framework to real-world scenarios.

\begin{acks}
The project is financially supported by the European Union (ERC, ``dynaBBO'', grant no.~101125586), by ANR project ANR-23-CE23-0035 Opt4DAC.
André Biedenkapp acknowledges funding through the research network ``Responsive and Scalable Learning for Robots Assisting Humans'' (ReScaLe) of the University of Freiburg. The ReScaLe project is funded by the Carl Zeiss Foundation. This work used the Cirrus UK National Tier-2 HPC Service at EPCC (http://www.cirrus.ac.uk) funded by the University of Edinburgh and EPSRC (EP/P020267/1). Tai Nguyen acknowledges funding from the St Andrews Global Doctoral Scholarship programme.

\end{acks}

\bibliographystyle{ACM-Reference-Format}
\bibliography{refs}


\begin{thebibliography}{63}


\ifx \showCODEN    \undefined \def \showCODEN     #1{\unskip}     \fi
\ifx \showDOI      \undefined \def \showDOI       #1{#1}\fi
\ifx \showISBNx    \undefined \def \showISBNx     #1{\unskip}     \fi
\ifx \showISBNxiii \undefined \def \showISBNxiii  #1{\unskip}     \fi
\ifx \showISSN     \undefined \def \showISSN      #1{\unskip}     \fi
\ifx \showLCCN     \undefined \def \showLCCN      #1{\unskip}     \fi
\ifx \shownote     \undefined \def \shownote      #1{#1}          \fi
\ifx \showarticletitle \undefined \def \showarticletitle #1{#1}   \fi
\ifx \showURL      \undefined \def \showURL       {\relax}        \fi
\providecommand\bibfield[2]{#2}
\providecommand\bibinfo[2]{#2}
\providecommand\natexlab[1]{#1}
\providecommand\showeprint[2][]{arXiv:#2}

\bibitem[Adriaensen et~al\mbox{.}(2022)]%
        {adriaensen2022automated}
\bibfield{author}{\bibinfo{person}{Steven Adriaensen}, \bibinfo{person}{Andr{\'e} Biedenkapp}, \bibinfo{person}{Gresa Shala}, \bibinfo{person}{Noor Awad}, \bibinfo{person}{Theresa Eimer}, \bibinfo{person}{Marius Lindauer}, {and} \bibinfo{person}{Frank Hutter}.} \bibinfo{year}{2022}\natexlab{}.
\newblock \showarticletitle{Automated dynamic algorithm configuration}.
\newblock \bibinfo{journal}{\emph{Journal of Artificial Intelligence Research}}  \bibinfo{volume}{75} (\bibinfo{year}{2022}), \bibinfo{pages}{1633--1699}.
\newblock


\bibitem[Aine et~al\mbox{.}(2008)]%
        {aine-jasoc08a}
\bibfield{author}{\bibinfo{person}{Sandip Aine}, \bibinfo{person}{Rajeev Kumar}, {and} \bibinfo{person}{P.P. Chakrabarti}.} \bibinfo{year}{2008}\natexlab{}.
\newblock \showarticletitle{Adaptive parameter control of evolutionary algorithms to improve quality-time trade-off}.
\newblock \bibinfo{journal}{\emph{Applied Soft Computing}} \bibinfo{volume}{9}, \bibinfo{number}{2} (\bibinfo{year}{2008}), \bibinfo{pages}{527--540}.
\newblock
\urldef\tempurl%
\url{https://doi.org/10.1016/j.asoc.2008.07.001}
\showURL{%
\tempurl}


\bibitem[Andersson et~al\mbox{.}(2016)]%
        {andersson-gecco16b}
\bibfield{author}{\bibinfo{person}{Martin Andersson}, \bibinfo{person}{Sunith Bandaru}, {and} \bibinfo{person}{Amos~HC Ng}.} \bibinfo{year}{2016}\natexlab{}.
\newblock \showarticletitle{Tuning of Multiple Parameter Sets in Evolutionary Algorithms}. In \bibinfo{booktitle}{\emph{Proceedings of the Genetic and Evolutionary Computation Conference 2016}} (2016-01-01). \bibinfo{pages}{533--540}.
\newblock
\urldef\tempurl%
\url{https://dl.acm.org/doi/10.1145/2908812.2908899}
\showURL{%
\tempurl}


\bibitem[Github(2025)]%
        {source}
\bibfield{author}{\bibinfo{person}{Github}.} \bibinfo{year}{2025}\natexlab{}.
\newblock \bibinfo{howpublished}{\url{https://github.com/taindp98/OneMax-DAC.git}}.
\newblock


\bibitem[Bellemare et~al\mbox{.}(2016)]%
        {bellemare2016unifying}
\bibfield{author}{\bibinfo{person}{Marc Bellemare}, \bibinfo{person}{Sriram Srinivasan}, \bibinfo{person}{Georg Ostrovski}, \bibinfo{person}{Tom Schaul}, \bibinfo{person}{David Saxton}, {and} \bibinfo{person}{Remi Munos}.} \bibinfo{year}{2016}\natexlab{}.
\newblock \showarticletitle{Unifying count-based exploration and intrinsic motivation}.
\newblock \bibinfo{journal}{\emph{Advances in neural information processing systems}}  \bibinfo{volume}{29} (\bibinfo{year}{2016}).
\newblock


\bibitem[Bellemare et~al\mbox{.}(2020)]%
        {loon}
\bibfield{author}{\bibinfo{person}{Marc~G. Bellemare}, \bibinfo{person}{Salvatore Candido}, \bibinfo{person}{Pablo~Samuel Castro}, \bibinfo{person}{Jun Gong}, \bibinfo{person}{Marlos~C. Machado}, \bibinfo{person}{Subhodeep Moitra}, \bibinfo{person}{Sameera~S. Ponda}, {and} \bibinfo{person}{Ziyu Wang}.} \bibinfo{year}{2020}\natexlab{}.
\newblock \showarticletitle{Autonomous navigation of stratospheric balloons using reinforcement learning}.
\newblock \bibinfo{journal}{\emph{Nat.}} \bibinfo{volume}{588}, \bibinfo{number}{7836} (\bibinfo{year}{2020}), \bibinfo{pages}{77--82}.
\newblock


\bibitem[Bellman(1957)]%
        {MDP}
\bibfield{author}{\bibinfo{person}{R. Bellman}.} \bibinfo{year}{1957}\natexlab{}.
\newblock \showarticletitle{A Markovian decision process}.
\newblock \bibinfo{journal}{\emph{Journal of Mathematics and Mechanics}} (\bibinfo{year}{1957}), \bibinfo{pages}{679--684}.
\newblock


\bibitem[Benjamins et~al\mbox{.}(2024)]%
        {benjamins2024instance}
\bibfield{author}{\bibinfo{person}{Carolin Benjamins}, \bibinfo{person}{Gjorgjina Cenikj}, \bibinfo{person}{Ana Nikolikj}, \bibinfo{person}{Aditya Mohan}, \bibinfo{person}{Tome Eftimov}, {and} \bibinfo{person}{Marius Lindauer}.} \bibinfo{year}{2024}\natexlab{}.
\newblock \showarticletitle{Instance selection for dynamic algorithm configuration with reinforcement learning: Improving generalization}. In \bibinfo{booktitle}{\emph{Proceedings of the Genetic and Evolutionary Computation Conference Companion}}. \bibinfo{pages}{563--566}.
\newblock


\bibitem[Biedenkapp et~al\mbox{.}(2020)]%
        {biedenkapp2020dynamic}
\bibfield{author}{\bibinfo{person}{Andr{\'e} Biedenkapp}, \bibinfo{person}{H~Furkan Bozkurt}, \bibinfo{person}{Theresa Eimer}, \bibinfo{person}{Frank Hutter}, {and} \bibinfo{person}{Marius Lindauer}.} \bibinfo{year}{2020}\natexlab{}.
\newblock \showarticletitle{Dynamic algorithm configuration: Foundation of a new meta-algorithmic framework}.
\newblock In \bibinfo{booktitle}{\emph{ECAI 2020}}. \bibinfo{publisher}{IOS Press}, \bibinfo{pages}{427--434}.
\newblock


\bibitem[Biedenkapp et~al\mbox{.}(2022)]%
        {biedenkapp2022theory}
\bibfield{author}{\bibinfo{person}{Andr{\'e} Biedenkapp}, \bibinfo{person}{Nguyen Dang}, \bibinfo{person}{Martin~S Krejca}, \bibinfo{person}{Frank Hutter}, {and} \bibinfo{person}{Carola Doerr}.} \bibinfo{year}{2022}\natexlab{}.
\newblock \showarticletitle{Theory-inspired parameter control benchmarks for dynamic algorithm configuration}. In \bibinfo{booktitle}{\emph{Proceedings of the Genetic and Evolutionary Computation Conference}}. \bibinfo{pages}{766--775}.
\newblock


\bibitem[Bordne et~al\mbox{.}(2024)]%
        {bordne-automlws24a}
\bibfield{author}{\bibinfo{person}{Philipp Bordne}, \bibinfo{person}{M.~Asif Hasan}, \bibinfo{person}{Eddie Bergman}, \bibinfo{person}{Noor Awad}, {and} \bibinfo{person}{André Biedenkapp}.} \bibinfo{year}{2024}\natexlab{}.
\newblock \showarticletitle{CANDID DAC: Leveraging Coupled Action Dimensions with Importance Differences in DAC}. In \bibinfo{booktitle}{\emph{Proceedings of the Third International Conference on Automated Machine Learning (AutoML 2024), Workshop Track}}.
\newblock


\bibitem[Brafman and Tennenholtz(2002)]%
        {brafman2002r}
\bibfield{author}{\bibinfo{person}{Ronen~I Brafman} {and} \bibinfo{person}{Moshe Tennenholtz}.} \bibinfo{year}{2002}\natexlab{}.
\newblock \showarticletitle{R-max-a general polynomial time algorithm for near-optimal reinforcement learning}.
\newblock \bibinfo{journal}{\emph{Journal of Machine Learning Research}} \bibinfo{volume}{3}, \bibinfo{number}{Oct} (\bibinfo{year}{2002}), \bibinfo{pages}{213--231}.
\newblock


\bibitem[Burda et~al\mbox{.}(2018)]%
        {burda2018large}
\bibfield{author}{\bibinfo{person}{Yuri Burda}, \bibinfo{person}{Harri Edwards}, \bibinfo{person}{Deepak Pathak}, \bibinfo{person}{Amos Storkey}, \bibinfo{person}{Trevor Darrell}, {and} \bibinfo{person}{Alexei~A Efros}.} \bibinfo{year}{2018}\natexlab{}.
\newblock \showarticletitle{Large-scale study of curiosity-driven learning}.
\newblock \bibinfo{journal}{\emph{arXiv preprint arXiv:1808.04355}} (\bibinfo{year}{2018}).
\newblock


\bibitem[Burke et~al\mbox{.}(2013)]%
        {BurkeGHKOOQ13}
\bibfield{author}{\bibinfo{person}{Edmund~K. Burke}, \bibinfo{person}{Michel Gendreau}, \bibinfo{person}{Matthew~R. Hyde}, \bibinfo{person}{Graham Kendall}, \bibinfo{person}{Gabriela Ochoa}, \bibinfo{person}{Ender {\"{O}}zcan}, {and} \bibinfo{person}{Rong Qu}.} \bibinfo{year}{2013}\natexlab{}.
\newblock \showarticletitle{Hyper-heuristics: a survey of the state of the art}.
\newblock \bibinfo{journal}{\emph{J. Oper. Res. Soc.}} \bibinfo{volume}{64}, \bibinfo{number}{12} (\bibinfo{year}{2013}), \bibinfo{pages}{1695--1724}.
\newblock
\urldef\tempurl%
\url{https://doi.org/10.1057/jors.2013.71}
\showDOI{\tempurl}


\bibitem[Chen et~al\mbox{.}(2023)]%
        {chen2023using}
\bibfield{author}{\bibinfo{person}{Deyao Chen}, \bibinfo{person}{Maxim Buzdalov}, \bibinfo{person}{Carola Doerr}, {and} \bibinfo{person}{Nguyen Dang}.} \bibinfo{year}{2023}\natexlab{}.
\newblock \showarticletitle{Using automated algorithm configuration for parameter control}. In \bibinfo{booktitle}{\emph{Proceedings of the 17th ACM/SIGEVO Conference on Foundations of Genetic Algorithms}}. \bibinfo{pages}{38--49}.
\newblock


\bibitem[Choshen et~al\mbox{.}(2018)]%
        {choshen2018dora}
\bibfield{author}{\bibinfo{person}{Leshem Choshen}, \bibinfo{person}{Lior Fox}, {and} \bibinfo{person}{Yonatan Loewenstein}.} \bibinfo{year}{2018}\natexlab{}.
\newblock \showarticletitle{Dora the explorer: Directed outreaching reinforcement action-selection}.
\newblock \bibinfo{journal}{\emph{arXiv preprint arXiv:1804.04012}} (\bibinfo{year}{2018}).
\newblock


\bibitem[Degrave et~al\mbox{.}(2022)]%
        {degrave-nature22a}
\bibfield{author}{\bibinfo{person}{Jonas Degrave}, \bibinfo{person}{Federico Felici}, \bibinfo{person}{Jonas Buchli}, \bibinfo{person}{Michael Neunert}, \bibinfo{person}{Brendan~D. Tracey}, \bibinfo{person}{Francesco Carpanese}, \bibinfo{person}{Timo Ewalds}, \bibinfo{person}{Roland Hafner}, \bibinfo{person}{Abbas Abdolmaleki}, \bibinfo{person}{Diego de Las~Casas}, \bibinfo{person}{Craig Donner}, \bibinfo{person}{Leslie Fritz}, \bibinfo{person}{Cristian Galperti}, \bibinfo{person}{Andrea Huber}, \bibinfo{person}{James Keeling}, \bibinfo{person}{Maria Tsimpoukelli}, \bibinfo{person}{Jackie Kay}, \bibinfo{person}{Antoine Merle}, \bibinfo{person}{Jean{-}Marc Moret}, \bibinfo{person}{Seb Noury}, \bibinfo{person}{Federico Pesamosca}, \bibinfo{person}{David Pfau}, \bibinfo{person}{Olivier Sauter}, \bibinfo{person}{Cristian Sommariva}, \bibinfo{person}{Stefano Coda}, \bibinfo{person}{Basil Duval}, \bibinfo{person}{Ambrogio Fasoli}, \bibinfo{person}{Pushmeet Kohli}, \bibinfo{person}{Koray Kavukcuoglu},
  \bibinfo{person}{Demis Hassabis}, {and} \bibinfo{person}{Martin~A. Riedmiller}.} \bibinfo{year}{2022}\natexlab{}.
\newblock \showarticletitle{Magnetic control of tokamak plasmas through deep reinforcement learning}.
\newblock \bibinfo{journal}{\emph{Nat.}} \bibinfo{volume}{602}, \bibinfo{number}{7897} (\bibinfo{year}{2022}), \bibinfo{pages}{414--419}.
\newblock


\bibitem[Dey et~al\mbox{.}(2024)]%
        {dey2024continual}
\bibfield{author}{\bibinfo{person}{Sheelabhadra Dey}, \bibinfo{person}{James Ault}, {and} \bibinfo{person}{Guni Sharon}.} \bibinfo{year}{2024}\natexlab{}.
\newblock \showarticletitle{Continual optimistic initialization for value-based reinforcement learning}. In \bibinfo{booktitle}{\emph{Proceedings of the 23rd International Conference on Autonomous Agents and Multiagent Systems}}. \bibinfo{pages}{453--462}.
\newblock


\bibitem[Doerr and Doerr(2015)]%
        {doerr2015optimal}
\bibfield{author}{\bibinfo{person}{Benjamin Doerr} {and} \bibinfo{person}{Carola Doerr}.} \bibinfo{year}{2015}\natexlab{}.
\newblock \showarticletitle{Optimal parameter choices through self-adjustment: Applying the 1/5-th rule in discrete settings}. In \bibinfo{booktitle}{\emph{Proceedings of the 2015 Annual Conference on Genetic and Evolutionary Computation}}. \bibinfo{pages}{1335--1342}.
\newblock


\bibitem[Doerr and Doerr(2018)]%
        {doerr2018optimal}
\bibfield{author}{\bibinfo{person}{Benjamin Doerr} {and} \bibinfo{person}{Carola Doerr}.} \bibinfo{year}{2018}\natexlab{}.
\newblock \showarticletitle{Optimal static and self-adjusting parameter choices for the (1+($\lambda$, $\lambda$))(1+($\lambda$, $\lambda$)) genetic algorithm}.
\newblock \bibinfo{journal}{\emph{Algorithmica}}  \bibinfo{volume}{80} (\bibinfo{year}{2018}), \bibinfo{pages}{1658--1709}.
\newblock


\bibitem[Doerr et~al\mbox{.}(2015)]%
        {doerr2015black}
\bibfield{author}{\bibinfo{person}{Benjamin Doerr}, \bibinfo{person}{Carola Doerr}, {and} \bibinfo{person}{Franziska Ebel}.} \bibinfo{year}{2015}\natexlab{}.
\newblock \showarticletitle{From black-box complexity to designing new genetic algorithms}.
\newblock \bibinfo{journal}{\emph{Theoretical Computer Science}}  \bibinfo{volume}{567} (\bibinfo{year}{2015}), \bibinfo{pages}{87--104}.
\newblock


\bibitem[Eimer et~al\mbox{.}(2021)]%
        {eimer-ijcai21}
\bibfield{author}{\bibinfo{person}{T. Eimer}, \bibinfo{person}{A. Biedenkapp}, \bibinfo{person}{M. Reimer}, \bibinfo{person}{S. Adriaensen}, \bibinfo{person}{F. Hutter}, {and} \bibinfo{person}{M. Lindauer}.} \bibinfo{year}{2021}\natexlab{}.
\newblock \showarticletitle{DACBench: A Benchmark Library for Dynamic Algorithm Configuration}. In \bibinfo{booktitle}{\emph{Proceedings of the Thirtieth International Joint Conference on Artificial Intelligence ({IJCAI}'21)}}. \bibinfo{publisher}{ijcai.org}.
\newblock


\bibitem[Even-Dar and Mansour(2001)]%
        {even2001convergence}
\bibfield{author}{\bibinfo{person}{Eyal Even-Dar} {and} \bibinfo{person}{Yishay Mansour}.} \bibinfo{year}{2001}\natexlab{}.
\newblock \showarticletitle{Convergence of optimistic and incremental Q-learning}.
\newblock \bibinfo{journal}{\emph{Advances in neural information processing systems}}  \bibinfo{volume}{14} (\bibinfo{year}{2001}).
\newblock


\bibitem[Forbes et~al\mbox{.}(2024)]%
        {10.5555/3635637.3662910}
\bibfield{author}{\bibinfo{person}{Grant~C. Forbes}, \bibinfo{person}{Nitish Gupta}, \bibinfo{person}{Leonardo Villalobos-Arias}, \bibinfo{person}{Colin~M. Potts}, \bibinfo{person}{Arnav Jhala}, {and} \bibinfo{person}{David~L. Roberts}.} \bibinfo{year}{2024}\natexlab{}.
\newblock \showarticletitle{Potential-Based Reward Shaping for Intrinsic Motivation}. In \bibinfo{booktitle}{\emph{Proceedings of the 23rd International Conference on Autonomous Agents and Multiagent Systems}} (Auckland, New Zealand) \emph{(\bibinfo{series}{AAMAS '24})}. \bibinfo{publisher}{International Foundation for Autonomous Agents and Multiagent Systems}, \bibinfo{address}{Richland, SC}, \bibinfo{pages}{589–597}.
\newblock
\showISBNx{9798400704864}


\bibitem[Haarnoja et~al\mbox{.}(2018)]%
        {sac}
\bibfield{author}{\bibinfo{person}{Tuomas Haarnoja}, \bibinfo{person}{Aurick Zhou}, \bibinfo{person}{Kristian Hartikainen}, \bibinfo{person}{George Tucker}, \bibinfo{person}{Sehoon Ha}, \bibinfo{person}{Jie Tan}, \bibinfo{person}{Vikash Kumar}, \bibinfo{person}{Henry Zhu}, \bibinfo{person}{Abhishek Gupta}, \bibinfo{person}{Pieter Abbeel}, {and} \bibinfo{person}{Sergey Levine}.} \bibinfo{year}{2018}\natexlab{}.
\newblock \showarticletitle{Soft Actor-Critic Algorithms and Applications}.
\newblock \bibinfo{journal}{\emph{CoRR}}  \bibinfo{volume}{abs/1812.05905} (\bibinfo{year}{2018}).
\newblock


\bibitem[Hallak et~al\mbox{.}(2015)]%
        {hallak-corr15}
\bibfield{author}{\bibinfo{person}{Assaf Hallak}, \bibinfo{person}{Dotan~Di Castro}, {and} \bibinfo{person}{Shie Mannor}.} \bibinfo{year}{2015}\natexlab{}.
\newblock \showarticletitle{Contextual Markov Decision Processes}.
\newblock \bibinfo{journal}{\emph{CoRR}}  \bibinfo{volume}{abs/1502.02259} (\bibinfo{year}{2015}).
\newblock
\urldef\tempurl%
\url{http://arxiv.org/abs/1502.02259}
\showURL{%
\tempurl}


\bibitem[Hansen(2006)]%
        {hansen2006cma}
\bibfield{author}{\bibinfo{person}{Nikolaus Hansen}.} \bibinfo{year}{2006}\natexlab{}.
\newblock \showarticletitle{The {CMA} evolution strategy: a comparing review}.
\newblock \bibinfo{journal}{\emph{Towards a new evolutionary computation: Advances in the estimation of distribution algorithms}} (\bibinfo{year}{2006}), \bibinfo{pages}{75--102}.
\newblock


\bibitem[Hutter et~al\mbox{.}(2009)]%
        {hutter-jair09a}
\bibfield{author}{\bibinfo{person}{Frank Hutter}, \bibinfo{person}{Holger~H. Hoos}, \bibinfo{person}{Kevin Leyton{-}Brown}, {and} \bibinfo{person}{Thomas St{\"{u}}tzle}.} \bibinfo{year}{2009}\natexlab{}.
\newblock \showarticletitle{Param{ILS}: An Automatic Algorithm Configuration Framework}.
\newblock   \bibinfo{volume}{36} (\bibinfo{year}{2009}), \bibinfo{pages}{267--306}.
\newblock


\bibitem[Karafotias et~al\mbox{.}(2012)]%
        {KarafotiasSE12}
\bibfield{author}{\bibinfo{person}{Giorgos Karafotias}, \bibinfo{person}{Selmar~K. Smit}, {and} \bibinfo{person}{A.~E. Eiben}.} \bibinfo{year}{2012}\natexlab{}.
\newblock \showarticletitle{A Generic Approach to Parameter Control}. In \bibinfo{booktitle}{\emph{Proc. of Applications of Evolutionary Computation (EvoApplications'12)}} \emph{(\bibinfo{series}{LNCS}, Vol.~\bibinfo{volume}{7248})}. \bibinfo{publisher}{Springer}, \bibinfo{pages}{366--375}.
\newblock
\urldef\tempurl%
\url{https://doi.org/10.1007/978-3-642-29178-4\_37}
\showDOI{\tempurl}


\bibitem[Kaufmann et~al\mbox{.}(2023)]%
        {droneracing}
\bibfield{author}{\bibinfo{person}{Elia Kaufmann}, \bibinfo{person}{Leonard Bauersfeld}, \bibinfo{person}{Antonio Loquercio}, \bibinfo{person}{Matthias M{\"{u}}ller}, \bibinfo{person}{Vladlen Koltun}, {and} \bibinfo{person}{Davide Scaramuzza}.} \bibinfo{year}{2023}\natexlab{}.
\newblock \showarticletitle{Champion-level drone racing using deep reinforcement learning}.
\newblock \bibinfo{journal}{\emph{Nat.}} \bibinfo{volume}{620}, \bibinfo{number}{7976} (\bibinfo{year}{2023}), \bibinfo{pages}{982--987}.
\newblock


\bibitem[Kee et~al\mbox{.}(2001)]%
        {KeeAdaptiveGA}
\bibfield{author}{\bibinfo{person}{Eric Kee}, \bibinfo{person}{Sarah Airey}, {and} \bibinfo{person}{Walling Cyre}.} \bibinfo{year}{2001}\natexlab{}.
\newblock \showarticletitle{An Adaptive Genetic Algorithm}. In \bibinfo{booktitle}{\emph{Proc. of Genetic and Evolutionary Computation Conference (GECCO'01)}}. \bibinfo{publisher}{Morgan Kaufmann}, \bibinfo{pages}{391–397}.
\newblock
\showISBNx{1558607749}
\urldef\tempurl%
\url{https://doi.org/10.5555/2955239.2955303}
\showDOI{\tempurl}


\bibitem[Kingma and Ba(2015)]%
        {adam}
\bibfield{author}{\bibinfo{person}{Diederik~P. Kingma} {and} \bibinfo{person}{Jimmy Ba}.} \bibinfo{year}{2015}\natexlab{}.
\newblock \showarticletitle{Adam: {A} Method for Stochastic Optimization}. In \bibinfo{booktitle}{\emph{Proceedings of the 3rd International Conference on Learning Representations, ({ICLR}'15)}}, \bibfield{editor}{\bibinfo{person}{Yoshua Bengio} {and} \bibinfo{person}{Yann LeCun}} (Eds.).
\newblock


\bibitem[Lagoudakis et~al\mbox{.}(2000)]%
        {lagoudakis2000algorithm}
\bibfield{author}{\bibinfo{person}{Michail~G Lagoudakis}, \bibinfo{person}{Michael~L Littman}, {et~al\mbox{.}}} \bibinfo{year}{2000}\natexlab{}.
\newblock \showarticletitle{Algorithm Selection using Reinforcement Learning.}. In \bibinfo{booktitle}{\emph{ICML}}. \bibinfo{pages}{511--518}.
\newblock


\bibitem[Laud(2004)]%
        {laud2004theory}
\bibfield{author}{\bibinfo{person}{Adam~Daniel Laud}.} \bibinfo{year}{2004}\natexlab{}.
\newblock \bibinfo{booktitle}{\emph{Theory and application of reward shaping in reinforcement learning}}.
\newblock \bibinfo{publisher}{University of Illinois at Urbana-Champaign}.
\newblock


\bibitem[Levine and Abbeel(2014)]%
        {levine-neurips14}
\bibfield{author}{\bibinfo{person}{S. Levine} {and} \bibinfo{person}{P. Abbeel}.} \bibinfo{year}{2014}\natexlab{}.
\newblock \showarticletitle{Learning Neural Network Policies with Guided Policy Search under Unknown Dynamics}. In \bibinfo{booktitle}{\emph{Proceedings of the 28th International Conference on Advances in Neural Information Processing Systems ({N}eur{IPS}'14)}}, \bibfield{editor}{\bibinfo{person}{Z.~Ghahramani}, \bibinfo{person}{M.~Welling}, \bibinfo{person}{C.~Cortes}, \bibinfo{person}{N.~Lawrence}, {and} \bibinfo{person}{K.~Weinberger}} (Eds.). \bibinfo{pages}{1071--1079}.
\newblock


\bibitem[Lillicrap(2015)]%
        {lillicrap2015continuous}
\bibfield{author}{\bibinfo{person}{TP Lillicrap}.} \bibinfo{year}{2015}\natexlab{}.
\newblock \showarticletitle{Continuous control with deep reinforcement learning}.
\newblock \bibinfo{journal}{\emph{arXiv preprint arXiv:1509.02971}} (\bibinfo{year}{2015}).
\newblock


\bibitem[Lillicrap et~al\mbox{.}(2016)]%
        {ddpg-soft-update}
\bibfield{author}{\bibinfo{person}{Timothy~P. Lillicrap}, \bibinfo{person}{Jonathan~J. Hunt}, \bibinfo{person}{Alexander Pritzel}, \bibinfo{person}{Nicolas Heess}, \bibinfo{person}{Tom Erez}, \bibinfo{person}{Yuval Tassa}, \bibinfo{person}{David Silver}, {and} \bibinfo{person}{Daan Wierstra}.} \bibinfo{year}{2016}\natexlab{}.
\newblock \showarticletitle{Continuous control with deep reinforcement learning}. In \bibinfo{booktitle}{\emph{4th International Conference on Learning Representations, {ICLR} 2016, San Juan, Puerto Rico, May 2-4, 2016, Conference Track Proceedings}}, \bibfield{editor}{\bibinfo{person}{Yoshua Bengio} {and} \bibinfo{person}{Yann LeCun}} (Eds.).
\newblock
\urldef\tempurl%
\url{http://arxiv.org/abs/1509.02971}
\showURL{%
\tempurl}


\bibitem[Lindauer et~al\mbox{.}(2022)]%
        {SMAC3}
\bibfield{author}{\bibinfo{person}{Marius Lindauer}, \bibinfo{person}{Katharina Eggensperger}, \bibinfo{person}{Matthias Feurer}, \bibinfo{person}{Andr{\'{e}} Biedenkapp}, \bibinfo{person}{Difan Deng}, \bibinfo{person}{Carolin Benjamins}, \bibinfo{person}{Tim Ruhkopf}, \bibinfo{person}{Ren{\'{e}} Sass}, {and} \bibinfo{person}{Frank Hutter}.} \bibinfo{year}{2022}\natexlab{}.
\newblock \showarticletitle{{SMAC3:} {A} Versatile Bayesian Optimization Package for Hyperparameter Optimization}.
\newblock \bibinfo{journal}{\emph{J. Mach. Learn. Res.}}  \bibinfo{volume}{23} (\bibinfo{year}{2022}), \bibinfo{pages}{54:1--54:9}.
\newblock
\urldef\tempurl%
\url{https://jmlr.org/papers/v23/21-0888.html}
\showURL{%
\tempurl}


\bibitem[L{\'o}pez-Ib{\'a}{\~n}ez et~al\mbox{.}(2016)]%
        {lopez2016irace}
\bibfield{author}{\bibinfo{person}{Manuel L{\'o}pez-Ib{\'a}{\~n}ez}, \bibinfo{person}{J{\'e}r{\'e}mie Dubois-Lacoste}, \bibinfo{person}{Leslie~P{\'e}rez C{\'a}ceres}, \bibinfo{person}{Mauro Birattari}, {and} \bibinfo{person}{Thomas St{\"u}tzle}.} \bibinfo{year}{2016}\natexlab{}.
\newblock \showarticletitle{The irace package: Iterated racing for automatic algorithm configuration}.
\newblock \bibinfo{journal}{\emph{Operations Research Perspectives}}  \bibinfo{volume}{3} (\bibinfo{year}{2016}), \bibinfo{pages}{43--58}.
\newblock


\bibitem[Ma et~al\mbox{.}({[n.\,d.]})]%
        {mareward}
\bibfield{author}{\bibinfo{person}{Haozhe Ma}, \bibinfo{person}{Kuankuan Sima}, \bibinfo{person}{Thanh~Vinh Vo}, \bibinfo{person}{Di Fu}, {and} \bibinfo{person}{Tze-Yun Leong}.} \bibinfo{year}{[n.\,d.]}\natexlab{}.
\newblock \showarticletitle{Reward Shaping for Reinforcement Learning with An Assistant Reward Agent}. In \bibinfo{booktitle}{\emph{Forty-first International Conference on Machine Learning}}.
\newblock


\bibitem[Mnih(2013)]%
        {mnih2013playing}
\bibfield{author}{\bibinfo{person}{Volodymyr Mnih}.} \bibinfo{year}{2013}\natexlab{}.
\newblock \showarticletitle{Playing atari with deep reinforcement learning}.
\newblock \bibinfo{journal}{\emph{arXiv preprint arXiv:1312.5602}} (\bibinfo{year}{2013}).
\newblock


\bibitem[Ng et~al\mbox{.}(1999)]%
        {ng1999theory}
\bibfield{author}{\bibinfo{person}{Andrew~Y Ng}, \bibinfo{person}{Daishi Harada}, {and} \bibinfo{person}{Stuart Russell}.} \bibinfo{year}{1999}\natexlab{}.
\newblock \showarticletitle{Theory and application to reward shaping}. In \bibinfo{booktitle}{\emph{Proceedings of the Sixteenth International Conference on Machine Learning}}.
\newblock


\bibitem[Osband et~al\mbox{.}(2016)]%
        {osband2016deep}
\bibfield{author}{\bibinfo{person}{Ian Osband}, \bibinfo{person}{Charles Blundell}, \bibinfo{person}{Alexander Pritzel}, {and} \bibinfo{person}{Benjamin Van~Roy}.} \bibinfo{year}{2016}\natexlab{}.
\newblock \showarticletitle{Deep exploration via bootstrapped DQN}.
\newblock \bibinfo{journal}{\emph{Advances in neural information processing systems}}  \bibinfo{volume}{29} (\bibinfo{year}{2016}).
\newblock


\bibitem[Parker-Holder et~al\mbox{.}(2022)]%
        {parker-holder-jair22a}
\bibfield{author}{\bibinfo{person}{Jack Parker-Holder}, \bibinfo{person}{Raghu Rajan}, \bibinfo{person}{Xingyou Song}, \bibinfo{person}{André Biedenkapp}, \bibinfo{person}{Yingjie Miao}, \bibinfo{person}{Theresa Eimer}, \bibinfo{person}{Baohe Zhang}, \bibinfo{person}{Vu Nguyen}, \bibinfo{person}{Roberto Calandra}, \bibinfo{person}{Aleksandra Faust}, \bibinfo{person}{Frank Hutter}, {and} \bibinfo{person}{Marius Lindauer}.} \bibinfo{year}{2022}\natexlab{}.
\newblock \showarticletitle{Automated Reinforcement Learning (AutoRL): A Survey and Open Problems}.
\newblock \bibinfo{journal}{\emph{Journal of Artificial Intelligence Research (JAIR)}}  \bibinfo{volume}{74} (\bibinfo{year}{2022}), \bibinfo{pages}{517--568}.
\newblock
\urldef\tempurl%
\url{https://doi.org/10.1613/jair.1.13596}
\showDOI{\tempurl}


\bibitem[Pettinger and Everson(2002)]%
        {pettinger-gecco02b}
\bibfield{author}{\bibinfo{person}{J Pettinger} {and} \bibinfo{person}{R Everson}.} \bibinfo{year}{2002}\natexlab{}.
\newblock \showarticletitle{Controlling Genetic Algorithms with Reinforcement Learning}. In \bibinfo{booktitle}{\emph{Proceedings of the 4th Annual Conference on Genetic and Evolutionary Computation}} (2002-01-01). \bibinfo{pages}{692--692}.
\newblock
\urldef\tempurl%
\url{https://dl.acm.org/doi/10.5555/2955491.2955607}
\showURL{%
\tempurl}


\bibitem[Randl{\o}v and Alstr{\o}m(1998)]%
        {randlov1998learning}
\bibfield{author}{\bibinfo{person}{Jette Randl{\o}v} {and} \bibinfo{person}{Preben Alstr{\o}m}.} \bibinfo{year}{1998}\natexlab{}.
\newblock \showarticletitle{Learning to Drive a Bicycle Using Reinforcement Learning and Shaping.}. In \bibinfo{booktitle}{\emph{ICML}}, Vol.~\bibinfo{volume}{98}. \bibinfo{pages}{463--471}.
\newblock


\bibitem[Sakurai et~al\mbox{.}(2010)]%
        {sakurai-sitis10ab}
\bibfield{author}{\bibinfo{person}{Y Sakurai}, \bibinfo{person}{K Takada}, \bibinfo{person}{T Kawabe}, {and} \bibinfo{person}{S Tsuruta}.} \bibinfo{year}{2010}\natexlab{}.
\newblock \showarticletitle{A Method to Control Parameters of Evolutionary Algorithms by Using Reinforcement Learning}. In \bibinfo{booktitle}{\emph{Proceedings of Sixth International Conference on Signal-Image Technology and Internet-Based Systems (SITIS)}} (2010-01-01), \bibfield{editor}{\bibinfo{person}{K~Y é}, \bibinfo{person}{A~Dipanda}, {and} \bibinfo{person}{R~Chbeir}} (Eds.). \bibinfo{publisher}{IEEE Computer Society}, \bibinfo{pages}{74--79}.
\newblock
\urldef\tempurl%
\url{https://ieeexplore.ieee.org/document/5714532}
\showURL{%
\tempurl}


\bibitem[Schede et~al\mbox{.}(2022)]%
        {schede2022survey}
\bibfield{author}{\bibinfo{person}{Elias Schede}, \bibinfo{person}{Jasmin Brandt}, \bibinfo{person}{Alexander Tornede}, \bibinfo{person}{Marcel Wever}, \bibinfo{person}{Viktor Bengs}, \bibinfo{person}{Eyke H{\"u}llermeier}, {and} \bibinfo{person}{Kevin Tierney}.} \bibinfo{year}{2022}\natexlab{}.
\newblock \showarticletitle{A survey of methods for automated algorithm configuration}.
\newblock \bibinfo{journal}{\emph{Journal of Artificial Intelligence Research}}  \bibinfo{volume}{75} (\bibinfo{year}{2022}), \bibinfo{pages}{425--487}.
\newblock


\bibitem[Shala et~al\mbox{.}(2020)]%
        {shala2020learning}
\bibfield{author}{\bibinfo{person}{Gresa Shala}, \bibinfo{person}{Andr{\'e} Biedenkapp}, \bibinfo{person}{Noor Awad}, \bibinfo{person}{Steven Adriaensen}, \bibinfo{person}{Marius Lindauer}, {and} \bibinfo{person}{Frank Hutter}.} \bibinfo{year}{2020}\natexlab{}.
\newblock \showarticletitle{Learning step-size adaptation in CMA-ES}. In \bibinfo{booktitle}{\emph{Parallel Problem Solving from Nature--PPSN XVI: 16th International Conference, PPSN 2020, Leiden, The Netherlands, September 5-9, 2020, Proceedings, Part I 16}}. Springer, \bibinfo{pages}{691--706}.
\newblock


\bibitem[Sharma et~al\mbox{.}(2019)]%
        {sharma2019deep}
\bibfield{author}{\bibinfo{person}{Mudita Sharma}, \bibinfo{person}{Alexandros Komninos}, \bibinfo{person}{Manuel L{\'o}pez-Ib{\'a}{\~n}ez}, {and} \bibinfo{person}{Dimitar Kazakov}.} \bibinfo{year}{2019}\natexlab{}.
\newblock \showarticletitle{Deep reinforcement learning based parameter control in differential evolution}. In \bibinfo{booktitle}{\emph{Proceedings of the genetic and evolutionary computation conference}}. \bibinfo{pages}{709--717}.
\newblock


\bibitem[Silver et~al\mbox{.}(2017)]%
        {silver2017mastering}
\bibfield{author}{\bibinfo{person}{David Silver}, \bibinfo{person}{Julian Schrittwieser}, \bibinfo{person}{Karen Simonyan}, \bibinfo{person}{Ioannis Antonoglou}, \bibinfo{person}{Aja Huang}, \bibinfo{person}{Arthur Guez}, \bibinfo{person}{Thomas Hubert}, \bibinfo{person}{Lucas Baker}, \bibinfo{person}{Matthew Lai}, \bibinfo{person}{Adrian Bolton}, {et~al\mbox{.}}} \bibinfo{year}{2017}\natexlab{}.
\newblock \showarticletitle{Mastering the game of go without human knowledge}.
\newblock \bibinfo{journal}{\emph{nature}} \bibinfo{volume}{550}, \bibinfo{number}{7676} (\bibinfo{year}{2017}), \bibinfo{pages}{354--359}.
\newblock


\bibitem[Strehl and Littman(2004)]%
        {strehl2004empirical}
\bibfield{author}{\bibinfo{person}{Alexander~L Strehl} {and} \bibinfo{person}{Michael~L Littman}.} \bibinfo{year}{2004}\natexlab{}.
\newblock \showarticletitle{An empirical evaluation of interval estimation for markov decision processes}. In \bibinfo{booktitle}{\emph{16th IEEE International Conference on Tools with Artificial Intelligence}}. IEEE, \bibinfo{pages}{128--135}.
\newblock


\bibitem[Sullivan et~al\mbox{.}(2024)]%
        {sullivan2024reward}
\bibfield{author}{\bibinfo{person}{Ryan Sullivan}, \bibinfo{person}{Akarsh Kumar}, \bibinfo{person}{Shengyi Huang}, \bibinfo{person}{John Dickerson}, {and} \bibinfo{person}{Joseph Suarez}.} \bibinfo{year}{2024}\natexlab{}.
\newblock \showarticletitle{Reward scale robustness for proximal policy optimization via DreamerV3 tricks}.
\newblock \bibinfo{journal}{\emph{Advances in Neural Information Processing Systems}}  \bibinfo{volume}{36} (\bibinfo{year}{2024}).
\newblock


\bibitem[Sun et~al\mbox{.}(2022)]%
        {sun2022exploit}
\bibfield{author}{\bibinfo{person}{Hao Sun}, \bibinfo{person}{Lei Han}, \bibinfo{person}{Rui Yang}, \bibinfo{person}{Xiaoteng Ma}, \bibinfo{person}{Jian Guo}, {and} \bibinfo{person}{Bolei Zhou}.} \bibinfo{year}{2022}\natexlab{}.
\newblock \showarticletitle{Exploit reward shifting in value-based deep-rl: Optimistic curiosity-based exploration and conservative exploitation via linear reward shaping}.
\newblock \bibinfo{journal}{\emph{Advances in neural information processing systems}}  \bibinfo{volume}{35} (\bibinfo{year}{2022}), \bibinfo{pages}{37719--37734}.
\newblock


\bibitem[Sutton and Barto(1998)]%
        {712192}
\bibfield{author}{\bibinfo{person}{R.S. Sutton} {and} \bibinfo{person}{A.G. Barto}.} \bibinfo{year}{1998}\natexlab{}.
\newblock \showarticletitle{Reinforcement Learning: An Introduction}.
\newblock \bibinfo{journal}{\emph{IEEE Transactions on Neural Networks}} \bibinfo{volume}{9}, \bibinfo{number}{5} (\bibinfo{year}{1998}), \bibinfo{pages}{1054--1054}.
\newblock
\urldef\tempurl%
\url{https://doi.org/10.1109/TNN.1998.712192}
\showDOI{\tempurl}


\bibitem[Sutton(1988)]%
        {sutton1988learning}
\bibfield{author}{\bibinfo{person}{Richard~S Sutton}.} \bibinfo{year}{1988}\natexlab{}.
\newblock \showarticletitle{Learning to predict by the methods of temporal differences}.
\newblock \bibinfo{journal}{\emph{Machine learning}}  \bibinfo{volume}{3} (\bibinfo{year}{1988}), \bibinfo{pages}{9--44}.
\newblock


\bibitem[Szita and L{\H{o}}rincz(2008)]%
        {szita2008many}
\bibfield{author}{\bibinfo{person}{Istv{\'a}n Szita} {and} \bibinfo{person}{Andr{\'a}s L{\H{o}}rincz}.} \bibinfo{year}{2008}\natexlab{}.
\newblock \showarticletitle{The many faces of optimism: a unifying approach}. In \bibinfo{booktitle}{\emph{Proceedings of the 25th international conference on Machine learning}}. \bibinfo{pages}{1048--1055}.
\newblock


\bibitem[Van~Hasselt et~al\mbox{.}(2016a)]%
        {van2016deep}
\bibfield{author}{\bibinfo{person}{Hado Van~Hasselt}, \bibinfo{person}{Arthur Guez}, {and} \bibinfo{person}{David Silver}.} \bibinfo{year}{2016}\natexlab{a}.
\newblock \showarticletitle{Deep reinforcement learning with double q-learning}. In \bibinfo{booktitle}{\emph{Proceedings of the AAAI conference on artificial intelligence}}, Vol.~\bibinfo{volume}{30}.
\newblock


\bibitem[Van~Hasselt et~al\mbox{.}(2016b)]%
        {van2016learning}
\bibfield{author}{\bibinfo{person}{Hado~P Van~Hasselt}, \bibinfo{person}{Arthur Guez}, \bibinfo{person}{Matteo Hessel}, \bibinfo{person}{Volodymyr Mnih}, {and} \bibinfo{person}{David Silver}.} \bibinfo{year}{2016}\natexlab{b}.
\newblock \showarticletitle{Learning values across many orders of magnitude}.
\newblock \bibinfo{journal}{\emph{Advances in neural information processing systems}}  \bibinfo{volume}{29} (\bibinfo{year}{2016}).
\newblock


\bibitem[Vermetten et~al\mbox{.}(2019)]%
        {VermettenCMAdynAS}
\bibfield{author}{\bibinfo{person}{Diederick Vermetten}, \bibinfo{person}{Sander van Rijn}, \bibinfo{person}{Thomas B{\"{a}}ck}, {and} \bibinfo{person}{Carola Doerr}.} \bibinfo{year}{2019}\natexlab{}.
\newblock \showarticletitle{Online selection of {CMA-ES} variants}. In \bibinfo{booktitle}{\emph{Proc. of Genetic and Evolutionary Computation Conference (GECCO'19)}}. \bibinfo{publisher}{ACM}, \bibinfo{pages}{951--959}.
\newblock
\urldef\tempurl%
\url{https://doi.org/10.1145/3321707.3321803}
\showDOI{\tempurl}


\bibitem[Watkins and Dayan(1992)]%
        {watkins1992q}
\bibfield{author}{\bibinfo{person}{Christopher~JCH Watkins} {and} \bibinfo{person}{Peter Dayan}.} \bibinfo{year}{1992}\natexlab{}.
\newblock \showarticletitle{Q-learning}.
\newblock \bibinfo{journal}{\emph{Machine learning}}  \bibinfo{volume}{8} (\bibinfo{year}{1992}), \bibinfo{pages}{279--292}.
\newblock


\bibitem[Wurman et~al\mbox{.}(2022)]%
        {gtsophy}
\bibfield{author}{\bibinfo{person}{Peter~R. Wurman}, \bibinfo{person}{Samuel Barrett}, \bibinfo{person}{Kenta Kawamoto}, \bibinfo{person}{James MacGlashan}, \bibinfo{person}{Kaushik Subramanian}, \bibinfo{person}{Thomas~J. Walsh}, \bibinfo{person}{Roberto Capobianco}, \bibinfo{person}{Alisa Devlic}, \bibinfo{person}{Franziska Eckert}, \bibinfo{person}{Florian Fuchs}, \bibinfo{person}{Leilani Gilpin}, \bibinfo{person}{Piyush Khandelwal}, \bibinfo{person}{Varun~Raj Kompella}, \bibinfo{person}{HaoChih Lin}, \bibinfo{person}{Patrick MacAlpine}, \bibinfo{person}{Declan Oller}, \bibinfo{person}{Takuma Seno}, \bibinfo{person}{Craig Sherstan}, \bibinfo{person}{Michael~D. Thomure}, \bibinfo{person}{Houmehr Aghabozorgi}, \bibinfo{person}{Leon Barrett}, \bibinfo{person}{Rory Douglas}, \bibinfo{person}{Dion Whitehead}, \bibinfo{person}{Peter D{\"{u}}rr}, \bibinfo{person}{Peter Stone}, \bibinfo{person}{Michael Spranger}, {and} \bibinfo{person}{Hiroaki Kitano}.} \bibinfo{year}{2022}\natexlab{}.
\newblock \showarticletitle{Outracing champion Gran Turismo drivers with deep reinforcement learning}.
\newblock \bibinfo{journal}{\emph{Nat.}} \bibinfo{volume}{602}, \bibinfo{number}{7896} (\bibinfo{year}{2022}), \bibinfo{pages}{223--228}.
\newblock


\bibitem[Xue et~al\mbox{.}(2022)]%
        {madac}
\bibfield{author}{\bibinfo{person}{Ke Xue}, \bibinfo{person}{Jiacheng Xu}, \bibinfo{person}{Lei Yuan}, \bibinfo{person}{Miqing Li}, \bibinfo{person}{Chao Qian}, \bibinfo{person}{Zongzhang Zhang}, {and} \bibinfo{person}{Yang Yu}.} \bibinfo{year}{2022}\natexlab{}.
\newblock \showarticletitle{Multi-agent Dynamic Algorithm Configuration}. In \bibinfo{booktitle}{\emph{Advances in Neural Information Processing Systems 35: Annual Conference on Neural Information Processing Systems, {NeurIPS'22}}}, \bibfield{editor}{\bibinfo{person}{Sanmi Koyejo}, \bibinfo{person}{S.~Mohamed}, \bibinfo{person}{A.~Agarwal}, \bibinfo{person}{Danielle Belgrave}, \bibinfo{person}{K.~Cho}, {and} \bibinfo{person}{A.~Oh}} (Eds.).
\newblock


\end{thebibliography}

\appendix
\section{Reward Shifting}
\label{apsec:reward_shifting}
The theory of reward shaping was introduced in \cite{ng1999theory}, where the authors presented the concept of learning another MDP model, denoted as $\mathcal{M}'$, instead of the original $\mathcal{M}$. The new $\mathcal{M}'$ is defined as $(\mathcal{S}, \mathcal{A}, \mathcal{T}, \mathcal{R}')$, where $\mathcal{R}'=\mathcal{R}+F$ is the reward set in  $\mathcal{M}'$. The trained agent in $\mathcal{M'}$ would also receive a reward of $R(s,a,s') + F(s,a,s')$ when executing the action $a$ to transition from state $s$ to $s'$. They defined a potential-based shaping function following Theorem 1 in \cite{ng1999theory} $\mathcal{F}: \mathcal{S}\times\mathcal{A}\times\mathcal{S} \mapsto \mathbb{R}$ and a real-value function $\Phi: \mathcal{S} \mapsto \mathbb{R}$:
\begin{equation}
\label{eq:potential_based_shaping}
F(s,a,s') = \gamma \Phi(s') - \Phi(s),
\end{equation}

A transformation from the optimal action-value function $\mathcal{Q}_{\mathcal{M}}$ in $\mathcal{M}$ to $\mathcal{M}'$ satisfies the Bellman equation:
\begin{equation}
\mathcal{Q}^{*}_{\mathcal{M}'}(s,a) = \mathcal{Q}^{*}_{\mathcal{M}}(s,a) - \Phi(s)
\end{equation}
and the optimal policy for $\mathcal{M}'$:
\begin{equation}
\label{eq:optimize_m_prime}
\pi^{*}_{\mathcal{M}'}(s) = \argmax_{a \in \mathcal{A}} \mathcal{Q}^{*}_{\mathcal{M}'}(s,a) = \argmax_{a \in \mathcal{A}} \mathcal{Q}^{*}_{\mathcal{M}}(s,a) - \Phi(s)
\end{equation}

Sun et al. \cite{sun2022exploit} defined the potential-based function $F$ in ~\Cref{eq:potential_based_shaping} as a constant bias $F(s,a,s') = b$, thus $\mathcal{R}'=\mathcal{R}+b$ with $b \in \mathbb{R}$, the formula in  ~\Cref{eq:potential_based_shaping} simplifies to:
\begin{equation}
F(s,a,s') = \gamma \Phi(s') - \Phi(s) = (\gamma - 1) \phi
\end{equation}
in the case where $F$ is guaranteed to be a constant, thus the potential function $\Phi(s)$ must also be constant $\phi$, then:
\begin{equation}
\Phi(s) = \Phi(s') = \phi = \frac{b}{\gamma - 1}
\end{equation}
and the ~\Cref{eq:optimize_m_prime} becomes (see also the Remark 1 in \cite{sun2022exploit}):
\begin{align}
\label{eq:invariant_reward_shaping}
\pi^{*}_{\mathcal{M}'}(s) 
&= \argmax_{a \in \mathcal{A}} \mathcal{Q}^{*}_{\mathcal{M}'}(s, a) \notag \\
&= \argmax_{a \in \mathcal{A}} \mathcal{Q}^{*}_{\mathcal{M}}(s, a) - \frac{b}{\gamma - 1} \notag \\ 
&= \argmax_{a \in \mathcal{A}} \mathcal{Q}^{*}_{\mathcal{M}}(s, a) + \frac{b}{1 - \gamma}
\end{align}
as the additional bias $b$ that does not depend on the chosen action leads to maximizing the action-value function $\mathcal{Q}_{\mathcal{M}'}$ which is equivalent to maximizing the original $\mathcal{Q}_\mathcal{M}$. The constant $\left| \frac{b}{1-\gamma} \right|$ represents the difference between the altered and the original state.

\section{Policy Comparison}

\begin{figure}[H]
    \centering
    \begin{subfigure}{0.49\linewidth}
        \includegraphics[width=\linewidth, clip]{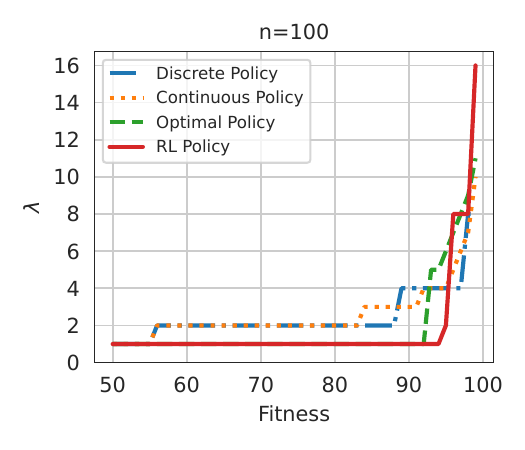}
    \end{subfigure}
    \begin{subfigure}{0.49\linewidth}
        \includegraphics[width=\linewidth, clip]{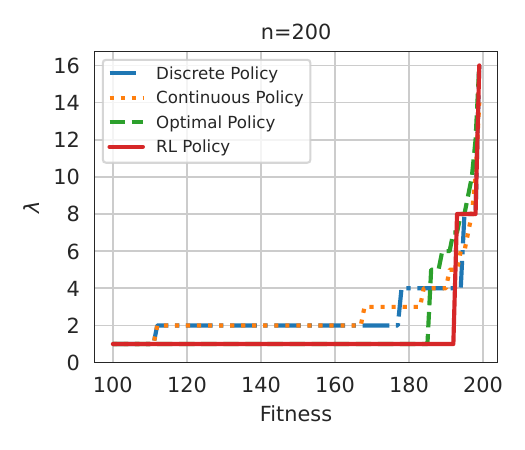}
    \end{subfigure}
    \begin{subfigure}{0.49\linewidth}
        \includegraphics[width=\linewidth, clip]{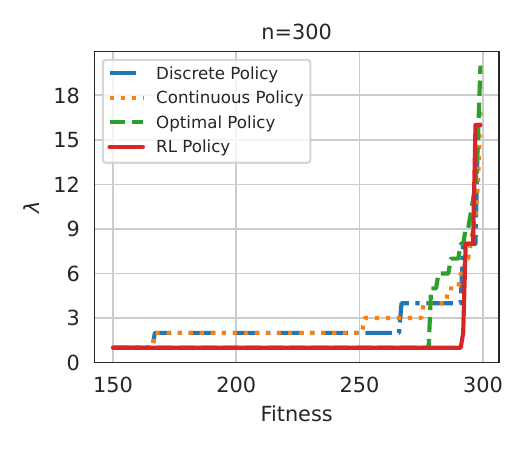}
    \end{subfigure}
    \begin{subfigure}{0.49\linewidth}
        \includegraphics[width=\linewidth, clip]{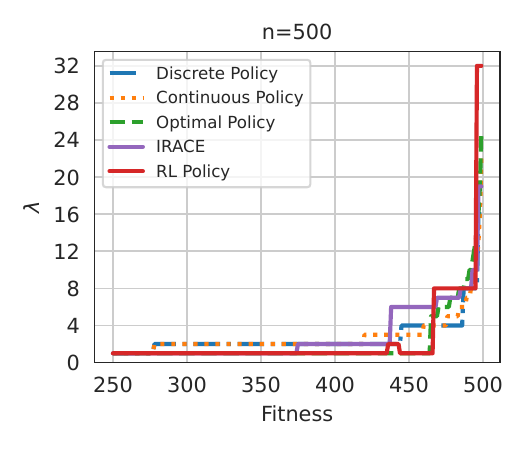}
    \end{subfigure}
    
    \begin{subfigure}{0.49\linewidth}
        \includegraphics[width=\linewidth, clip]{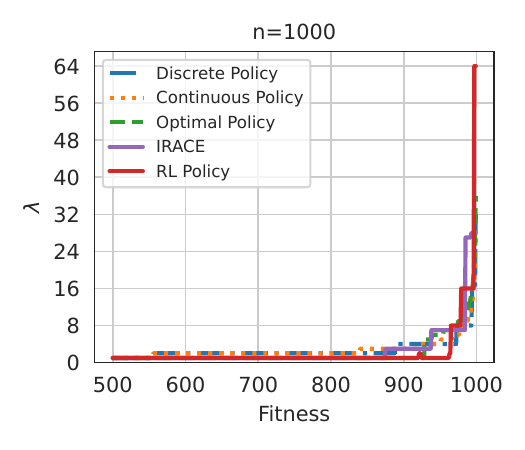}
    \end{subfigure}
    \begin{subfigure}{0.49\linewidth}
        \includegraphics[width=\linewidth, clip]{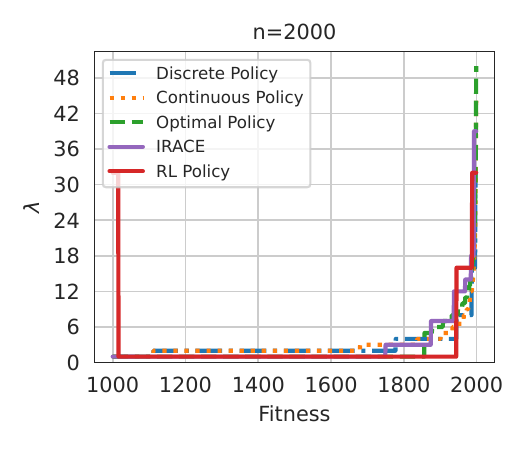}
    \end{subfigure}

    \caption{Policies of the theory-derived \( \pi_{\texttt{cont}} \), its discretized version \( \pi_{\texttt{disc}} \), the optimal policy \( \pi_{\texttt{opt}} \), \irace-based policies, and RL, which denotes our best-trained DDQN for \( f(x) \geq n/2 \).}
    \label{fig:policy_comparison}
\end{figure}

We present in~\Cref{fig:policy_comparison} the policies across 4 approaches for problem sizes $n \in \{100, 200, 300\}$ and an additional comparison with \irace for problem sizes $n \in \{500, 1000, 2000\}$. For RL, we present the best policies selected from the top-5 policies in the evaluation phase during training. These policies incorporate the reward shifting mechanism into the original reward function. 
Similar to~\cite{chen2023using}, we plot \(\lambda\) values only for \( f(x) \geq n/2 \), as this is the most relevant region.

\end{document}